\documentclass[article]{jss}
\usepackage{booktabs,amsmath,amsfonts,orcidlink,thumbpdf,lmodern}
\newtheorem{theorem}{Theorem}
\graphicspath{{Figures/}}
\shortcites{pedregosa2011}

\providecommand{\tightlist}{\setlength{\itemsep}{0pt}\setlength{\parskip}{0pt}}

\author{
  Philipp Bach~\orcidlink{0000-0002-7183-9239}\\University of Hamburg \And
  Malte S. Kurz\\Technical University\\ of Munich \And
  Victor Chernozhukov~\orcidlink{0000-0002-3250-6714}\\Massachusetts Institute\\of Technology \AND
  Martin Spindler~\orcidlink{0000-0002-1294-7782}\\University of Hamburg \And
  Sven Klaassen~\orcidlink{0009-0004-9080-0809}\\University of Hamburg
}
\Plainauthor{Philipp Bach, Malte S. Kurz, Victor Chernozhukov, Martin Spindler, Sven Klaassen}

\title{\pkg{DoubleML}: An Object-Oriented Implementation of Double Machine Learning in \proglang{R}}
\Plaintitle{DoubleML: An Object-Oriented Implementation of Double Machine Learning in R}
\Shorttitle{\pkg{DoubleML}: Double Machine Learning in \proglang{R}}

\Abstract{
  The \proglang{R}~package \pkg{DoubleML} implements the double/debiased
  machine learning framework of \citet{dml2018}. It provides
  functionalities to estimate parameters in causal models based on machine
  learning methods. The double machine learning framework consists of
  three key ingredients\text{:} Neyman orthogonality, high-quality machine
  learning estimation and sample splitting. Estimation of nuisance
  components can be performed by various state-of-the-art machine learning
  methods that are available in the \pkg{mlr3} ecosystem. \pkg{DoubleML}
  makes it possible to perform inference in a variety of causal models,
  including partially linear and interactive regression models and their
  extensions to instrumental variable estimation. The object-oriented
  implementation of \pkg{DoubleML} enables a high flexibility for the
  model specification and makes it easily extendable. This paper serves as
  an introduction to the double machine learning framework and the
  \proglang{R}~package \pkg{DoubleML}. In reproducible code examples with
  simulated and real data sets, we demonstrate how \pkg{DoubleML} users
  can perform valid inference based on machine learning methods.
}

\Plainkeywords{machine learning, causal inference, causal machine learning, R, mlr3, object orientation}

\Volume{108}
\Issue{3}
\Month{February}
\Year{2024}
\Submitdate{2021-06-25}
\Acceptdate{2023-04-14}
\DOI{10.18637/jss.v108.i03}

\Address{
  Philipp Bach, Martin Spindler, Sven Klaassen\\
  University of Hamburg\\
  Chair of Statistics\\
  Moorweidenstr. 18\\
  20148 Hamburg, Germany\\
  E-mail: \email{philipp.bach@uni-hamburg.de}, \email{martin.spindler@uni-hamburg.de},\\
  \phantom{E-mail: }\email{sven.klaassen@uni-hamburg.de}\\

  Malte S. Kurz\\
  Technical University of Munich\\
  TUM School of Management\\
  Arcisstr. 21\\
  80333 Munich, Germany\\
  E-mail: \email{malte.kurz@tum.de}\\

  Victor Chernozhukov\\
  Massachusetts Institute of Technology\\
  Department of Economics and Center for Statistics and Data Science\\
  50 Memorial Drive\\
  Cambridge, MA 02139, United States of America\\
  E-mail: \email{vchern@mit.edu}\\
}

\begin{document}

\vspace*{-0.5cm}

\section{Introduction} \label{intro}

Structural equation models provide a quintessential framework for
conducting causal inference in statistics, econometrics, machine
learning (ML), and other data sciences. The package \pkg{DoubleML} \citep{DoubleMLR}
for \proglang{R} \citep{R} implements partially linear and interactive structural
equation and treatment effect models with high-dimensional confounding
variables as considered in \citet{dml2018}. Estimation and tuning of the
machine learning models is based on the powerful functionalities
provided by the \pkg{mlr3} package and the \pkg{mlr3} ecosystem
\citep{mlr3}. A key ingredient of double machine learning (DML) models are
score functions identifying the estimates for the target parameter.
These functions play an essential role for valid inference with machine
learning methods because they have to satisfy a property called Neyman
orthogonality. With the score functions as key elements,
\pkg{DoubleML} implements double machine learning in a very general
way using object orientation based on the \pkg{R6} package
\citep{R6}. Currently, \pkg{DoubleML} implements the double /
debiased machine learning framework as established in \citet{dml2018}
for
\begin{itemize}
\tightlist
\item
  partially linear regression models (PLR),
\item
  partially linear instrumental variable regression models (PLIV),
\item
  interactive regression models (IRM), and,
\item
  interactive instrumental variable regression models (IIVM).
\end{itemize}
The object-oriented implementation of \pkg{DoubleML} is very
flexible. The model classes\linebreak `\code{DoubleMLPLR}', `\code{DoubleMLPLIV}',
`\code{DoubleMLIRM}' and `\code{DoubleIIVM}' implement the estimation of
the nuisance functions via machine learning methods and the computation
of the Neyman-orthogonal score function. All other functionalities are
implemented in the abstract base class `\code{DoubleML}', including
estimation of causal parameters, standard errors, $t$~tests,
confidence intervals, as well as valid simultaneous inference through
adjustments of $p$~values and estimation of joint confidence regions
based on a multiplier bootstrap procedure. In combination with the
estimation and tuning functionalities of \pkg{mlr3} and its
ecosystem, this object-oriented implementation enables a high
flexibility for the model specification in terms of
\begin{itemize}
\tightlist
\item
  the machine learning methods for estimation of the nuisance functions,
\item
  the resampling schemes,
\item
  the double machine learning algorithm, and,
\item
  the Neyman-orthogonal score functions.
\end{itemize}
It further can be readily extended regarding
\begin{itemize}
\tightlist
\item
  new model classes that come with Neyman-orthogonal score functions
  being linear in the target parameter,
\item
  alternative score functions via callables, and,
\item
  customized resampling schemes.
\end{itemize}
Several other \proglang{R}~packages for estimation of causal effects based on machine
learning methods exist for \proglang{R}. The packages \pkg{grf} \citep{grf} and \pkg{hdi}
\citep{hdi} implement alternative approaches to causal machine learning. \pkg{grf} implements generalized random forests \citep{grfpaper} and can be used for forest-based inference methods in different causal models including least-squares regression and estimation of treatment effects with and without instrumental variables.  \pkg{hdi} can be used for inference in high-dimensional models with a focus on lasso-based estimation and methods for simultaneous inference.

An alternative approach, which was developed before the double machine
learning framework, is the so-called targeted learning framework, and
its software implementations and ecosystem (\pkg{tlverse}). For an
overview and introduction to this approach and its implementations, we
refer to the extensive \pkg{tlverse} handbook \citep[][\url{https://tlverse.org/tlverse-handbook}]{tlmanual}. Relevant \proglang{R}~packages include \pkg{SuperLearner}
\citep{SuperLearner} for flexible estimation using machine learning and
\pkg{tmle} \citep{tmle} which implements estimation of causal
parameters using targeted maximum likelihood estimation (TMLE).
\pkg{sl3} \citep{sl3} and \pkg{tmle3} \citep{tmle3} are recent
extensions of the \pkg{tlverse} for object-oriented implementation of
machine learning algorithms and a unified interface for TMLE.

Previous implementations that are more closely related to the double machine learning framework of
\citet{dml2018} have been provided by the \proglang{R} packages \pkg{hdm} \citep{hdm}, \pkg{dmlmt}
\citep{knaus2018}, \pkg{causalDML} \citep{knaus2020}, \pkg{causalweight} \citep{causalweight} and \pkg{AIPW}
\citep{AIPW}. \pkg{hdm} offers lasso-based inference methods in a variety of high-dimensional causal models, including estimation of (local) average treatment effects and linear (instrumental variable) regression. The underlying theoretical framework for valid post-selection and post-regularization inference can be considered as a special case of the more generic DML framework of \citet{dml2018}. Similarly, \pkg{dmlt} \citep{knaus2018} provides methods of lasso-based inference on treatment effects of multi-valued treatment variables. \citet{knaus2020} focuses on a nonparametric treatment effect model, which is called the interactive regression model (IRM) in \cite{dml2018} and also referred to as augmented inverse probability weighting. We adapted the term IRM from \cite{dml2018} to denote this causal model and will refer to it accordingly in the following. The model is introduced in Section~\ref{irmmodel}. Allowing for additional learners that include generalized random forests, ridge and random forests, \pkg{causalDML} \citep{knaus2020} provides an implementation of the DML appraoch in an IRM in combination with recent methods for the analysis of heterogeneous treatment effects. Similarly, \pkg{causalweight} \citep{causalweight} focuses on the IRM and offers estimation methods for various causal quantities in this model as well as extensions thereof, including instrumental variable estimation, mediation analysis and sample selection approaches.
In line with \cite{dml2018}, we denote the instrumental variable extension of the IRM in the following as IIVM. The model is introduced in Section~\ref{iivmmodel}.
The \proglang{R}~package \pkg{AIPW}
\citep{AIPW} implements estimation of average treatment effects of a
binary treatment variable by augmented inverse probability weighting
based on machine learning algorithms and integrates well with the
\pkg{tlverse} ecosystem discussed later. In \proglang{Python} \citep{python}, \pkg{EconML} \citep{econml} offers an implementation of several causal machine learning approaches. The package does not exclusively build on the double machine learning framework by \citet{dml2018} and has a focus on heterogeneous effects.

In contrast to existing software packages, the \proglang{R}~package \pkg{DoubleML} is intended to be a general implementation of the double machine learning approach of \citet{dml2018}.
An introduction to the three key ingredients of the DML framework is provided in Section~\ref{basicidea}.
The package can be used to perform inference in basically any causal model that can be characterized in terms of the formal framework of \citet{dml2018}.
For example, it would be straightforward to extend \pkg{DoubleML} to
mediation analysis \citep{farbmacher2022}, sample selection models
\citep{bia2020} or difference-in-differences \citep{chang2020}. As we
will point out later, a key requirement for new model classes is a
Neyman-orthogonal score. The object-oriented implementation makes the package easily extendable in terms of the supported causal models and other features of DML. By building on the \pkg{mlr3} ecosystem estimation can be based on a rich collection of powerful ML methods available in \pkg{mlr3} \citep{mlr3}, \pkg{mlr3learners} \citep{mlr3learners} and \pkg{mlr3extralearners} \citep{mlr3extralearners}.
The package \pkg{DoubleML} is available
from the Comprehensive \proglang{R} Archive Network (CRAN) at \url{https://CRAN.R-project.org/package=DoubleML}.

We would like to mention that the \proglang{R}~package \pkg{DoubleML} was developed
together with a \proglang{Python} twin \citep{DoubleMLpython} that is based on
\pkg{scikit-learn} \citep{pedregosa2011}. The \proglang{Python}~package is also
available via
{GitHub} (\url{https://github.com/DoubleML/doubleml-for-py}),
the Python Package Index ({PyPI}, \url{https://pypi.org/project/DoubleML}),
and {conda-forge} (\url{https://anaconda.org/conda-forge/doubleml}).\linebreak
Moreover, \citet{doubleml_serverless} provides a serverless
implementation of the \proglang{Python}~module \pkg{DoubleML}.

The rest of the paper is structured as follows: In Section
\ref{getstarted}, we briefly demonstrate how to install the
\pkg{DoubleML} package and give a short motivating example to
illustrate the major idea behind the double machine learning approach.
Section~\ref{causalmodels} introduces the main causal model classes
implemented in \pkg{DoubleML}. Section~\ref{basicidea} shortly
summarizes the main ideas behind the double machine learning approach
and reviews the key ingredients required for valid inference based on
machine learning methods. Section~\ref{dmlinference} presents the main
steps and algorithms of the double machine learning procedure for
inference on one or multiple target parameters. Section
\ref{implementationdetails} provides more detailed insights on the
implemented classes and methods of \pkg{DoubleML}. Section
\ref{illustration} contains real-data and simulation examples for
estimation of causal parameters using the \pkg{DoubleML} package.
Additionally, this section provides a brief simulation study that
illustrates the validity of the implemented methods in finite samples.
Section~\ref{conclusion} concludes the paper. The code output that has
been suppressed in the main text and further information regarding the
simulations are presented in the appendix. To make the code examples
fully reproducible, the entire code is available in a supplementary
zip file for this paper, as well as at
\url{https://github.com/DoubleML/DoubleMLReplicationCode}. We would like to note that minor numerical differences might occur on other platforms when replicating code examples that involve random forest learners
(see Appendix~\ref{computationinfra} for more information on the infrastructure used).

\vspace*{-0.25cm}

\section{Getting started} \label{getstarted}

\vspace*{-0.25cm}

\subsection{Installation}

The latest CRAN release of \pkg{DoubleML} can be installed using the
command

\vspace*{-0.25cm}

\begin{CodeChunk}
\begin{CodeInput}
R> install.packages("DoubleML")
\end{CodeInput}
\end{CodeChunk}

\vspace*{-0.25cm}

Alternatively, the development version can be downloaded and installed
from the {GitHub}
(\url{https://github.com/DoubleML/doubleml-for-r})
repository using
the command
\citep[previous installation of the \pkg{remotes}~package is required,][]{remotes}

\vspace*{-0.25cm}

\begin{CodeChunk}
\begin{CodeInput}
R> remotes::install_github("DoubleML/doubleml-for-r")
\end{CodeInput}
\end{CodeChunk}

Among others, \pkg{DoubleML} depends on the \proglang{R}~package \pkg{R6} for
object oriented implementation, \pkg{data.table} \citep{datatable}
for the underlying data structure, as well as the packages \pkg{mlr3}
\citep{mlr3}, \pkg{mlr3learners} \citep{mlr3learners} and
\pkg{mlr3tuning} \citep{mlr3tuning} for estimation of machine
learning methods, model tuning and parameter handling. Moreover, the
underlying packages of the machine learning methods that are called in
\pkg{mlr3} or \pkg{mlr3learners} must be installed, for example
the packages \pkg{glmnet} for lasso estimation \citep{glmnet} or
\pkg{ranger} \citep{ranger} for random forests.

Load the package after completed installation.
\begin{CodeChunk}
\begin{CodeInput}
R> library("DoubleML")
\end{CodeInput}
\end{CodeChunk}

\subsection{A motivating example: Basics of double machine learning} \label{anexample}

In the following, we provide a brief summary of and motivation to double
machine learning methods and show how the corresponding methods provided
by the \pkg{DoubleML} package can be applied. The data generating
process (DGP) is based on the introductory example in \citet{dml2018}.
We consider a partially linear model: Our major interest is to estimate
the causal parameter \(\theta\) in the following regression equation
\begin{align*}
\begin{aligned}
y_i = \theta d_i + g_0(x_i) + \zeta_i, & &\zeta_i \sim \mathcal{N}(0,1),
\end{aligned}
\end{align*} with covariates \(x_i \sim \mathcal{N}(0, \Sigma)\), where
\(\Sigma\) is a matrix with entries \(\Sigma_{kj} = 0.7^{\lvert j-k\rvert}\). In
the following, the regression relationship between the treatment
variable \(d_i\) and the covariates \(x_i\) will play an important role
\begin{align*}
\begin{aligned}
d_i = m_0(x_i) + v_i, & &v_i \sim \mathcal{N}(0,1).\\
\end{aligned}
\end{align*} The nuisance functions \(m_0\) and \(g_0\) are given by
\begin{align*}
m_0(x_i) &=  x_{i,1} + \frac{1}{4} \frac{\exp(x_{i,3})}{1+\exp(x_{i,3})}, \\
g_0(x_i) &= \frac{\exp(x_{i,1})}{1+\exp(x_{i,1})} + \frac{1}{4} x_{i,3}.
\end{align*} We construct a setting with \(n=500\) observations and
\(p=20\) explanatory variables to demonstrate the use of the estimators
provided in \pkg{DoubleML}. Moreover, we set the true value of the
parameter \(\theta\) to \(\theta=0.5\). The corresponding data
generating process is implemented in the function
\code{make\_plr\_CCDHNR2018()}. We start by generating a realization
of a data set as a `\code{data.table}' object, which is subsequently
used to create an instance of the data backend of class
`\code{DoubleMLData}'.
\begin{CodeChunk}
\begin{CodeInput}
R> library("DoubleML")
R> alpha <- 0.5
R> n_obs <- 500
R> n_vars <- 20
R> set.seed(1234)
R> data_plr <- make_plr_CCDDHNR2018(alpha = alpha, n_obs = n_obs,
+    dim_x = n_vars, return_type = "data.table")
\end{CodeInput}
\end{CodeChunk}
The data backend implements the causal model: We specify that we perform
inference on the effect of the treatment variable \(d_i\) on the
dependent variable \(y_i\).
\begin{CodeChunk}
\begin{CodeInput}
R> obj_dml_data <- DoubleMLData$new(data_plr, y_col = "y", d_cols = "d")
\end{CodeInput}
\end{CodeChunk}
In the next step, we choose the machine learning method as an object of
class `\code{Learner}' from \pkg{mlr3}, \pkg{mlr3learners}
\citep{mlr3learners} or \pkg{mlr3extralearners}
\citep{mlr3extralearners}. As we will point out later, we have to
estimate two nuisance functions in order to perform valid inference in
the partially linear regression model. Hence, we have to specify two
learners.
Moreover, we split the sample into two folds used for
cross-fitting (\code{n\_folds\ =\ 2}) in our
illustrating examples for simplicity. Two-fold cross-fitting makes it
necessary to estimate the ML models only twice, i.e.,~once per fold,
which reduces the computational costs. In practice, it is generally
recommended to choose a larger number of folds, cf.~Remark 3. The
default for the number of folds is \code{n\_folds\ =\ 5}.

Load \pkg{mlr3} and \pkg{mlr3learners} packages and suppress output during estimation.
\begin{CodeChunk}
\begin{CodeInput}
R> library("mlr3")
R> library("mlr3learners")
R> lgr::get_logger("mlr3")$set_threshold("warn") 
\end{CodeInput}
\end{CodeChunk}
Initialize a random forests learner with specified parameters.
\begin{CodeChunk}
\begin{CodeInput}
R> ml_l <- lrn("regr.ranger", num.trees = 100, mtry = n_vars,
+    min.node.size = 2, max.depth = 5)
R> ml_m <- lrn("regr.ranger", num.trees = 100, mtry = n_vars,
+    min.node.size = 2, max.depth = 5)
R> ml_g <- lrn("regr.ranger", num.trees = 100, mtry = n_vars,
+    min.node.size = 2, max.depth = 5)
\end{CodeInput}
\end{CodeChunk}
Initialize a causal model object, here a PLR.
\begin{CodeChunk}
\begin{CodeInput}
R> doubleml_plr <- DoubleMLPLR$new(obj_dml_data,
+    ml_l, ml_m, ml_g, n_folds = 2, score = "IV-type")
\end{CodeInput}
\end{CodeChunk}
To estimate the causal effect of variable \(d_i\) on \(y_i\), we call
the \code{fit()} method. To summarize the estimation output, we call the \code{summary()} method.
\begin{CodeChunk}
\begin{CodeInput}
R> doubleml_plr$fit()
R> doubleml_plr$summary()
\end{CodeInput}
\begin{CodeOutput}
Estimates and significance testing of the effect of target variables
  Estimate. Std. Error t value Pr(>|t|)    
d   0.51457    0.04522   11.38   <2e-16 ***
---
Signif. codes:  0 '***' 0.001 '**' 0.01 '*' 0.05 '.' 0.1 ' ' 1
\end{CodeOutput}
\end{CodeChunk}
The output shows that the estimated coefficient is close to the true
parameter \(\theta=0.5\). Moreover, there is evidence to reject the null
hypotheses \(H_0: \theta=0\) at all common significance levels.

\section{Key causal models} \label{causalmodels}

\begin{figure}[b!]
\centering
\includegraphics[width=0.55\textwidth]{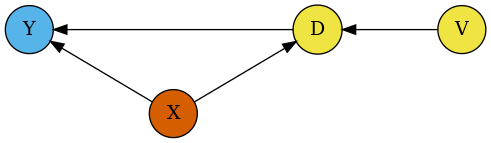}
\caption{%
  Causal diagram for PLR (Equation~\ref{plr1}--\ref{plr2}) and IRM
(Equation~\ref{irm1}--\ref{irm2}) under conditional exogeneity. Note that the
causal link between $D$ and $Y$ is one-directional. Identification of the causal
effect is confounded by $X$, and identification is achieved via $V$, which
captures variation in $D$ that is independent of $X$. Methods to estimate the
causal effect of $D$ must therefore approximately remove the effect of
high-dimensional $X$ on $Y$ and $D$.
}\label{dag1}
\end{figure}

\pkg{DoubleML} provides estimation of causal effects in four
different models: Partially linear regression models (PLR), partially
linear instrumental variable regression models (PLIV), interactive
regression models (IRM) and interactive instrumental variable regression
models (IIVM). We will shortly introduce these models.

\subsection{Partially linear regression model} \label{plrmodel}

Partially linear regression models (PLR), which encompass the standard
linear regression model, play an important role in data analysis
\citep{robinson1988}. Partially linear regression models take the form
\begin{align}
Y = D \theta_0 + g_0(X) + \zeta, \quad &\E(\zeta \mid D,X) = 0, \label{plr1}\\
D = m_0(X) + V, \quad &\E(V \mid X) = 0,   \label{plr2}
\end{align} where \(Y\) is the outcome variable and \(D\) is the policy
variable of interest. The high-dimensional vector \(X=(X_1, \dots ,X_p)\)
consists of other confounding covariates, and \(\zeta\) and \(V\) are
stochastic errors. Equation~\ref{plr1} is the equation of interest,
and \(\theta_0\) is the main regression coefficient that we would like
to infer. If \(D\) is conditionally exogenous (randomly assigned
conditional on \(X\)), \(\theta_0\) has the interpretation of a
structural or causal parameter. The causal diagram supporting such
interpretation is shown in Figure~\ref{dag1}. The second equation keeps
track of confounding, namely the dependence of \(D\) on
covariates/controls. The characteristics \(X\) affect the policy
variable \(D\) via the function \(m_0(X)\) and the outcome variable via
the function \(g_0(X)\). The partially linear model generalizes both
linear regression models, where functions \(g_0\) and \(m_0\) are linear
with respect to a collection of basis functions with respect to \(X\),
and approximately linear models.

An applied example from the economics literature is the analysis of the
causal effect of 401(k) pension plans on employees' net financial assets
by \citet{401k1} and \citet{401k2}. In these studies, which are based on
observational data, it is argued that eligibility for 401(k) pension
plans can be assumed to be conditionally exogenous, once it is
controlled for a set of confounders \(X\), for example income. Following
this argumentation and modelling approach, the estimate on \(\theta_0\)
as obtained by a PLR can be interpreted as the average treatment effect
of 401(k) eligibility on net financial assets. A reassessment and
summary of the 401(k) example is available in \citet{dml2018} as well as
on the \pkg{DoubleML} website (\url{https://docs.doubleml.org/stable/examples/R_double_ml_pension.html}).

\subsection{Partially linear instrumental variable regression model} \label{plivmodel}

We next consider the partially linear instrumental variable regression
model (PLIV)
\begin{align}
Y - D \theta_0 =  g_0(X) + \zeta, \quad &\E(\zeta \mid Z, X) = 0, \label{pliv1} \\
Z = m_0(X) + V,  \quad & \E(V \mid X) = 0. \label{pliv2}
\end{align} Note that this model is not a regression model unless
\(Z=D\). Model~\ref{pliv1}--\ref{pliv2} is a canonical model in
causal inference, going back to \citet{wright1928}, with the modern
difference being that \(g_0\) and \(m_0\) are nonlinear, potentially
complicated functions of high-dimensional \(X\). The idea of this model
is that there is a structural or causal relation between \(Y\) and
\(D\), captured by \(\theta_0\), and \(g_0(X) + \zeta\) is the
stochastic error, partly explained by covariates \(X\). \(V\) and
\(\zeta\) are stochastic errors that are not explained by \(X\). Since
\(Y\) and \(D\) are jointly determined, we need an external factor,
commonly referred to as an instrument, \(Z\), to create exogenous
variation in \(D\). Note that \(Z\) should affect \(D\). The \(X\) here
serve again as confounding factors, so we can think of variation in
\(Z\) as being exogenous only conditional on \(X\).

A simple contextual example is from biostatistics \citep{permutt1989},
where \(Y\) is a health outcome and \(D\) is an indicator of smoking.
Thus, \(\theta_0\) captures the effect of smoking on health. Health
outcome \(Y\) and smoking behavior \(D\) are treated as being jointly
determined. \(X\) represents patient characteristics, and \(Z\) could be
a doctor's advice not to smoke (or another behavioral treatment) that
may affect the outcome \(Y\) only through shifting the behavior \(D\),
conditional on characteristics \(X\).

\begin{figure}[t!]
  \centering
\includegraphics[width=0.41\textwidth]{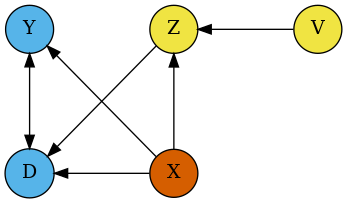}
\caption{%
  Causal diagram for PLIV (Equation~\ref{pliv1}--\ref{pliv2}) and IIVM
(Equation~\ref{iivm1}--\ref{iivm2}) under conditional exogeneity of $Z$. Note
that the causal link between $D$ and $Y$ is bi-directional, so an instrument $Z$
is needed for identification. Identification is achieved via $V$ that captures
variation in $Z$ that is independent of $X$. Equations~\ref{pliv1}
and~\ref{pliv2} do not model the dependence between $D$ and $X$ and $Z$, though
a necessary condition for identification is that $Z$ and $D$ are related after
conditioning on $X$. Methods to estimate the causal effect of $D$ must
approximately remove the effect of high-dimensional $X$ on $Y$, $D$, and $Z$.
Removing the confounding effect of $X$ is done implicitly by the proposed
procedure.
}\label{dag2}
\end{figure}

\subsection{Interactive regression model} \label{irmmodel}

We consider estimation of average treatment effects when treatment
effects are fully heterogeneous, i.e.,~the response curves under control
and treatment can be different nonparametric functions, and the
treatment variable is binary, \(D\in \{0, 1\}\). We consider vectors
\((Y, D, X)\) such that
\begin{align}
Y = g_0(D, X) + U, \quad & \E(U \mid X, D) = 0,  \label{irm1}\\
D = m_0(X) + V, \quad &\E(V \mid X) = 0. \label{irm2}
\end{align}
Since \(D\) is not additively separable, this model is more
general than the partially linear model for the case of binary \(D\). A
common target parameter of interest in this model is the average
treatment effect
(ATE).
\begin{align*}
\theta_0 = \E[g_0(1, X) - g_0(0,X)].
\end{align*}
Without unconfoundedness/conditional exogeneity, these quantities measure association, and could be referred to as average predictive effects (APE) and average predictive effect for the exposed (APEX). Inferential results for these objects would follow immediately from Theorem~\ref{theorem1}.

Another common target parameter is the average treatment
effect for the treated (ATTE)
\begin{align*}
\theta_0 = \E[g_0(1, X) - g_0(0,X) \mid D=1].
\end{align*} In business applications, the ATTE is often the main
interest, as it captures the treatment effect for those who have been
affected by the treatment. A difference of the ATTE from the ATE might
arise if the characteristics of the treated individuals differ from
those of the general population.

The confounding factors \(X\) affect the policy variable via the
propensity score \(m_0(X)\) and the outcome variable via the function
\(g_0(X)\). Both of these functions are unknown and potentially complex,
and we can employ ML methods to learn them.

Taking up the 401(k) example from Section~\ref{plrmodel}, the general
idea for identification of \(\theta_0\) using the IRM is similar. Once we are able to account for all confounding variables \(X\) in our
analysis, we can consistently estimate the causal parameter
\(\theta_0\). A difference to the PLR refers to assumptions on the
functional form of the main regression equation in~\ref{plr1} and
\ref{irm1}, respectively. Whereas it is assumed that the effect of \(D\)
on \(Y\) in the PLR model is additively separable, the IRM model comes
with less restrictive assumptions. For example, it is possible that
treatment effects are heterogeneous, i.e.,~vary across the population.

\subsection{Interactive instrumental variable model} \label{iivmmodel}

We consider estimation of the local average treatment effect (LATE) with
a binary treatment variable \(D\in\{0,1\}\), and a binary instrument,
\(Z\in\{0,1\}\). As before, \(Y\) denotes the outcome variable, and
\(X\) is the vector of covariates. In a setting where unobserved factors
drive the take-up of the treatment \(D\), the average treatment effect
is no longer identified. However, if a valid instrumental variable is
available that changes individuals' decision to take up the treatment,
it is possible to identify the LATE. The LATE measures the average
causal effect for the subgroup of compliers, i.e.,~those individuals who
receive the treatment only if the instrument takes value \(Z = 1\).
Hence, the LATE is of interest in many studies, where the treatment
assignment cannot be assumed to be conditionally independent. For a more
detailed treatment of the LATE and the key assumptions required for its
identification, we would like to refer to \citet{LATE}, \citet{mixtape}
and \citet{mostlyharmless}.

The structural equation for the IIVM is
\begin{align}
Y = \ell_0(D, X) + \zeta,  \quad & \E(\zeta \mid Z, X) = 0, \label{iivm1}\\
Z = m_0(X) + V, \quad &\E(V \mid X) = 0.\label{iivm2}
\end{align}
Consider the functions \(g_0\), \(r_0\), and \(m_0\), where
\(g_0\) maps the support of \((Z,X)\) to \(\mathbb{R}\) and \(r_0\) and
\(m_0\) map the support of \((Z,X)\) and \(X\) to
\((\epsilon, 1-\epsilon)\) for some \(\epsilon \in (0, 1/2)\), such that
\begin{align*}
Y = g_0(Z, X) + \nu, \quad &\E(\nu \mid Z, X) = 0,\\
D = r_0(Z, X) + U,  \quad & \E(U \mid Z, X) = 0,\\
Z = m_0(X) + V, \quad & \E(V \mid X) = 0.
\end{align*}
We are interested in estimating
\begin{align*} 
\theta_0 &= \frac{\E[g_0(1, X)] - \E[g_0(0,X)]}{\E[r_0(1, X)] - \E[r_0(0,X)]}.
\end{align*}
Under the well-known assumptions of \citet{LATE},
\(\theta_0\) is the LATE -- the average causal effect for compliers, in
other words, those observations that would have \(D=1\) if \(Z\) were
\(1\) and would have \(D=0\) if \(Z\) were \(0\).

In the smoking example from Section~\ref{plivmodel}, the setting is
similar to the section before, but now the binary treatment variable
(``smoking'') is endogenous and is instrumented by a binary instrument
variable \(Z\) (``doctor's advice''). In this example, the group of
compliers would comprise those individuals who quit smoking once their
doctor advises them to do so and would otherwise continue to smoke.
Similar to the comparison of the IRM model and the PLR model, the IIVM
model does not impose the assumptions of linearity and additive
separability that are maintained in the PLIV.

\section{Basic idea and key ingredients of double machine learning} \label{basicidea}

\subsection{Basic idea behind double machine learning for the PLR model} \label{basicideaplr}

\begin{figure}[t!]
  \centering
\includegraphics[width=0.8\textwidth, trim=15 20 0 0, clip]{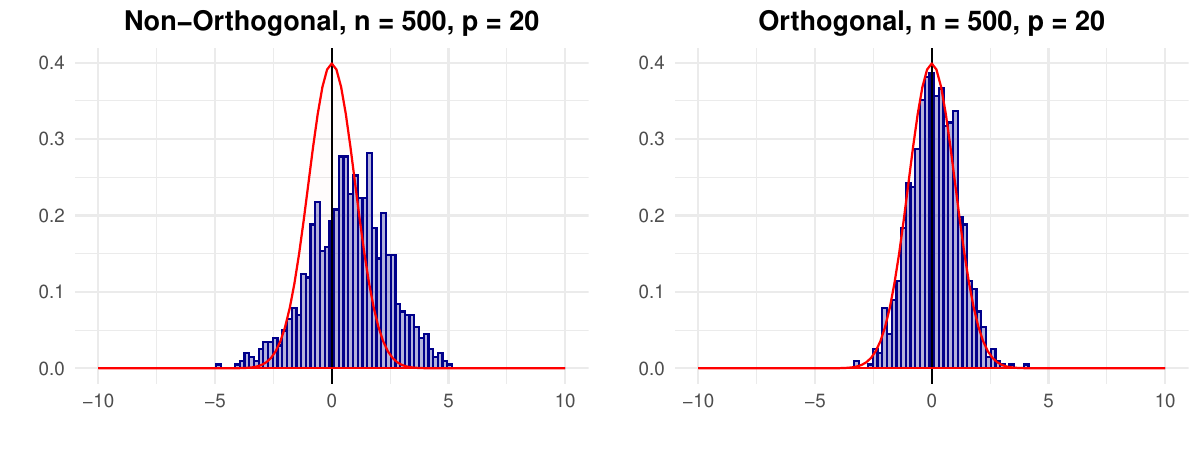}
\caption{%
  Performance of non-orthogonal and orthogonal estimators in a simulated data
example. Left panel: Histogram of the studentized naive estimator
$\hat{\theta}_{0}^{\text{naive}}$. $\hat{\theta}_{0}^{\text{naive}}$ is based on
estimation of $g_0$ and $m_0$ with random forests and a non-orthogonal score
function. Data sets are simulated according to the data generating process in
Section~\ref{anexample}. Data generation and estimation are repeated 1000 times.
Right panel: Histogram of the studentized DML estimator $\tilde{\theta}_0$.
$\tilde{\theta}_0$ is based on estimation of $g_0$ and $m_0$ with random forests
and an orthogonal score function provided in Equation~\ref{scorepartiallingout}.
Note that the simulated data sets and parameters of the random forest learners
are identical to those underlying the left panel.
}\label{failure1}
\end{figure}

Here we provide an intuitive discussion of how double machine learning
works in the first model, the partially linear regression model. Naive
application of machine learning methods directly to
Equations~\ref{plr1}--\ref{plr2} may have a very high bias. Indeed, it can be
shown that small biases in estimation of \(g_0\), which are unavoidable
in high-dimensional estimation, create a bias in the naive estimate of
the main effect, \(\hat{\theta}_{0}^{\text{naive}}\), which is sufficiently large
to cause failure of conventional inference. The left panel in Figure
\ref{failure1} illustrates this phenomenon. The histogram presents the
empirical distribution of the studentized estimator,
\(\hat{\theta}_{0}^{\text{naive}}\), as obtained in \(1000\) independent
repetitions of the data generating process presented in Section
\ref{anexample}. The functions \(g_0\) and \(m_0\) in the PLR model are
estimated with random forest learners and corresponding predictions are
then plugged into a non-orthogonal score function. The regularization
performed by the random forest learner leads to a bias in estimation of
\(g_0\) and \(m_0\). Due to non-orthogonality of the score, this
translates into a considerable bias of the main estimator
\(\hat{\theta}_{0}^{\text{naive}}\): The distribution of the studentized
estimator \(\hat{\theta}_{0}^{\text{naive}}\) is shifted to the right of the
origin and differs substantially from a normal distribution that would
be obtained if the regularization bias was negligible as shown by the
red curve.

The PLR model above can be rewritten in the following residualized form
\begin{align*}
W = V \theta_0 + \zeta,\quad & \E(\zeta \mid D, X) = 0, \\
W = (Y - \ell_0(X)), \quad & \ell_0(X) = \E[Y\mid X], \\
V = (D- m_0(X)), \quad & m_0(X) = \E[D\mid X].
\end{align*}
The variables \(W\) and \(V\) represent original variables
after taking out or \emph{partialling out} the effect of \(X\). Note
that \(\theta_0\) is identified from this equation if \(V\) has a
non-zero variance.

Given identification, double machine learning for a PLR proceeds as
follows
\begin{enumerate}
\item[(1)] Estimate $\ell_0$ and $m_0$ by $\hat{\ell}_0$ and $\hat{m}_0$, which amounts to solving the two problems of predicting $Y$ and $D$ using $X$, using any generic ML method, giving us estimated residuals
\begin{align*}
\hat{W} = Y - \hat{\ell}_0(X),
\end{align*}
and 
\begin{align*}
\hat{V} = D - \hat{m}_0(X).
\end{align*}
The residuals should be of a cross-validated form, as explained below in Algorithm 1 or 2, to avoid biases from overfitting. 
\item[(2)] Estimate $\theta_0$ by regressing the residual $\hat{W}$ on $\hat{V}$. Use the conventional inference for this regression estimator, ignoring the estimation error in the residuals. 
\end{enumerate}
The reason we work with this residualized form is that it eliminates the
bias arising from solving the prediction problems in stage (1). The
estimates \(\hat{\ell}_0\) and \(\hat{m}_0\) carry a regularization bias
due to having to solve prediction problems well in high-dimensions.
However, the nature of the estimating equation for \(\theta_0\) are such
that these biases are eliminated to the first order, as explained below.
This results in a high-quality low-bias estimator \(\tilde{\theta}_0\)
of \(\theta_0\), as illustrated in the right panel of Figure
\ref{failure1}. The estimator is adaptive in the sense that the first
stage estimation errors do not affect the second stage errors.

\subsection{Key ingredients of the double machine learning inference approach} \label{keyingredients}

Our goal is to construct high-quality point and interval estimators for
\(\theta_0\) when \(X\) is high-dimensional and we employ machine
learning methods to estimate the nuisance functions such as \(g_0\) and
\(m_0\). Example ML methods include lasso, random forests, boosted
trees, deep neural networks, and ensembles or aggregated versions of
these methods.

We shall use a method-of-moments estimator for \(\theta_0\) based upon
the empirical analog of the moment condition
\begin{align}
\E[\psi(W; \theta_0, \eta_0)] = 0, \label{score}
\end{align}
where we call \(\psi\) the score function,
\(W = (Y,D,X,Z)\), \(\theta_0\) is the parameter of interest, and
\(\eta\) denotes nuisance functions with population value \(\eta_0\).

\subsubsection{First key input: Neyman orthogonality}

The first key input of the inference procedure is using a
score function \(\psi(W; \theta; \eta)\) that satisfies~\ref{score},
with \(\theta_0\) being the unique solution, and that obeys the Neyman
orthogonality condition
\begin{align}
 \left.\partial_{\eta} \E[\psi(W; \theta_0, \eta)]\right|_{\eta=\eta_0}=0.   \label{neyman}
\end{align}
Neyman orthogonality~\ref{neyman} ensures that the moment condition
\ref{score} used to identify and estimate \(\theta_0\) is insensitive
to small pertubations of the nuisance function \(\eta\) around
\(\eta_0\). The derivative \(\partial_{\eta}\) denotes the pathwise
(Gateaux) derivative operator.

In general, it is important to distinguish whether machine learning
methods are used for prediction or in the context of statistical
inference. An accurate prediction rule for the nuisance parameters
\(\eta_0\) does not necessarily lead to a consistent estimator for the
causal parameter \(\theta_0\). Replacing the true value of \(\eta_0\) by
an ML estimator \(\hat{\eta}_0\) likely introduces a bias, for example,
due to heavy regularization in high-dimensional settings. If this bias
is not taken into account, the estimator \(\hat{\theta}_0\) will
generally be inconsistent and not have an asymptotically normal
distribution. Using a Neyman-orthogonal score makes estimation of the
causal parameter \(\theta_0\) robust against first order biases that arise
from regularization. The Neyman orthogonality property is responsible
for the adaptivity of the DML estimator -- namely, the approximate
distribution of \(\hat{\theta}_0\) will not depend on the fact that the
estimate \(\hat{\eta}_0\) contains error, if the latter is mild. Other
approaches, as targeted maximum likelihood and semiparametric sieves
estimation recognize this as well. For a more detailed treatment of
Neyman orthogonality we refer to \citet{dml2018}.

The right panel of Figure~\ref{failure1} presents the empirical
distribution of the studentized DML estimator \(\tilde{\theta}_0\) that
is based on an orthogonal score. Note that estimation is performed on
the identical simulated data sets and with the same machine learning
method as for the naive learner, which is displayed in the left panel.
The histogram of the studentized estimator \(\tilde{\theta}_0\)
illustrates the favorable performance of the double machine learning
estimator, which is based on an orthogonal score: The DML estimator is
robust to the bias that is generated by regularization. The estimator is
approximately unbiased, is concentrated around \(0\) and the
distribution is well-approximated by the normal distribution.

PLR score: In the PLR model, we can employ two alternative
  score functions. We will shortly indicate the option for
  initialization of a model object in \pkg{DoubleML} to clarify how
  each score can be implemented. Using the option
  \code{score\ =\ "partialling\ out"} leads to estimation of the score
  function
  \begin{align}  \label{scorepartiallingout}
  \begin{aligned}
  &\psi(W;\theta, \eta) := \left(Y - \ell(X) - \theta(D-m(X))\right) \left(D - m(X)\right),  \\
  & \eta = (\ell,m), \quad  \eta_0 = (\ell_0, m_0),
  \end{aligned}
  \end{align}
  where \(W=(Y,D,X)\) and \(\ell\) and \(m\) are
  \(P\)-square-integrable functions mapping the support of \(X\) to
  \(\mathbb{R}\), whose true values are given by
  \begin{align*}
  \ell_0(X) = \E[Y\mid X], \quad  m_0(X) = \E[D\mid X].
  \end{align*}
  Alternatively, it is possible to use the following score
  function for the PLR via the option \code{score\ =\ "IV-type"}
  \begin{align*}
  \begin{aligned}
  & \psi(W;\theta, \eta) := \left(Y-D\theta - g(X)\right) \left(D-m(X)\right), 
  & \eta = (g, m), \quad  \eta_0 = (g_0, m_0),
  \end{aligned}
  \end{align*}
  with \(g\) and \(m\) being \(P\)-square-integrable
  functions mapping the support of \(X\) to \(\mathbb{R}\) with values
  given by
  \begin{align*}
  g_0 = \E[Y-D\theta_0\mid X], \quad m_0(X) = \E[D\mid X].
  \end{align*}
  The scores above are Neyman-orthogonal by elementary
  calculations. Now, it is possible to see the connections to the
  residualized system of equations presented in Section
 ~\ref{basicideaplr}.

PLIV score: In the PLIV model, we can employ two alternative
  score functions. Using the option
  \code{score\ =\ "partialling\ out"} leads to estimation of the score
  function
\begin{align*}
  \begin{aligned}
  &\psi(W;\theta, \eta) := \left(Y - \ell(x) - \theta(D-r(X))\right) \left(Z - m(X) \right), \\
  & \eta = (\ell, m, r), \quad \eta_0 = (\ell_0, m_0, r_0),
  \end{aligned}
\end{align*}
where \(W=(Y,D,X,Z)\) and \(\ell\), \(m\), and \(r\) are
  \(P\)-square integrable functions mapping the support of \(X\) to
  \(\mathbb{R}\), whose true values are given by \begin{align*}
  \ell_0(X) = \E[Y\mid X], \quad r_0(X) = \E[D\mid X], \quad m_0(X) = \E[Z\mid X].
  \end{align*} Alternatively, it is possible to use the following score
  function for the PLIV via the option \code{score\ =\ "IV-type"}
  \begin{align*}
  \begin{aligned}
  & \psi(W;\theta, \eta) := \left(Y-D\theta - g(X)\right) \left(Z-m(X)\right), 
  & \eta = (g, m), \quad  \eta_0 = (g_0, m_0),
  \end{aligned}
  \end{align*}
  with \(g\) and \(m\) being \(P\)-square-integrable
  functions mapping the support of \(X\) to \(\mathbb{R}\) with values
  given by \begin{align*}
  g_0 = \E[Y-D\theta_0\mid X], \quad m_0(X) = \E[Z\mid X].
  \end{align*}

IRM score: For estimation of the ATE parameter of the IRM
model, we employ the score (\code{score\ =\ "ATE"})
\begin{align*}
  \begin{aligned}
  & \psi(W;\theta, \eta) := \left(g(1,X) - g(0,X) \right) + \frac{D(Y-g(1,X))}{m(X)} - \frac{(1-D)(Y-g(0,X))}{1-m(X)} - \theta, \\
  & \eta = (g,m), \quad \eta_0 = (g_0, m_0),
  \end{aligned}
\end{align*}
where \(W=(Y,D,X)\) and \(g\) and \(m\) map the support of
  \((D,X)\) to \(\mathbb{R}\) and the support of \(X\) to
  \((\epsilon, 1-\epsilon)\), respectively, for some
  \(\epsilon \in (0, 1/2)\), whose true values are given by
  \begin{align*}
  g_0(D,X)=\E[Y\mid D,X], \quad m_0(x)=\Prob[D=1\mid X].
  \end{align*}
  This orthogonal score is based on the influence function
  for the mean for missing data from \citet{robins1995}. For estimation
  of the ATTE parameter in the IRM, we use the score
  (\code{score\ =\ "ATTE"})
  \begin{align*}
  \begin{aligned}
  &\psi(W;\theta, \eta) := \frac{D(Y-g(0,X))}{p} - \frac{m(X)(1-D)(Y-g(0,X))}{p(1-m(x))} - \frac{D}{p}\theta,\\
  & \eta = (g, m, p), \quad \eta_0 = (g_0, m_0, p_0),
  \end{aligned}
  \end{align*}
  where \(p_0=\Prob(D=1)\). Note that this score does
  not require estimating \(g_0(1,X)\).

IIVM score: To estimate the LATE parameter in the IIVM, we
will use the score (\code{score\ =\ "LATE"})
\begin{align*}
  \begin{aligned}
  \psi := & g(1,X) - g(0,X) + \frac{Z(Y-g(1,X))}{m(X)} - \frac{(1-Z)(Y-g(0,X))}{1-m(X)} \\
  & - \left( r(1,x) - r(0,X) + \frac{Z(D-r(1,x)}{m(X)} - \frac{(1-Z)(D-r(0,X)}{1-m(X)} \right) \times \theta,\\
  & \eta = (g, m, r), \quad \eta_0 = (g_0, m_0, r_0),
  \end{aligned}
\end{align*}
where \(W=(Y,D,X,Z)\) and the nuisance parameter
  \(\eta=(g, m, r)\) consists of \(P\)-square integrable functions
  \(g\), \(m\), and \(r\), with \(g\) mapping the support of \((Z,X)\)
  to \(\mathbb{R}\) and \(m\) and \(r\), respectively, mapping the
  support of \((Z,X)\) and \(X\) to \((\epsilon, 1-\epsilon)\) for some
  \(\epsilon \in (0,1/2)\).

\subsubsection{Second key input: High-quality machine learning methods}
 
The second key input is the use of high-quality machine
learning estimators for the nuisance parameters.

For instance, in the PLR model with \code{score = "IV-type"}, we need to have access to consistent
estimators of \(g_0\) and \(m_0\) with respect to the \(L^2(P)\) norm
\(\lVert \cdot \lVert_{P,2}\), such that
\begin{align*}
\lVert \hat{m}_0 - m_0 \lVert_{P,2} + \lVert \hat{\ell}_0 - \ell_0 \lVert_{P,2} \le o(N^{-1/4}).
\end{align*}
In the PLIV model, the sufficient condition is
\begin{align*}
\lVert \hat{r}_0 - r_0 \lVert_{P,2} + \lVert \hat{m}_0 - m_0 \lVert_{P,2} + \lVert \hat{\ell}_0 - \ell_0 \lVert_{P,2} \le o(N^{-1/4}).
\end{align*}
These conditions are plausible for many ML methods.
Different structured assumptions on \(\eta_0\) lead to the use of
different machine-learning tools for estimating \(\eta_0\) as listed in
\citet[pp.~22--23]{dml2018}:
\begin{enumerate}
\def\labelenumi{\arabic{enumi}.}
\tightlist
\item
  The assumption of approximate or exact sparsity for \(\eta_0\) with
  respect to some set of regressors, known as dictionary in computer
  science, calls for the use of sparsity-based machine learning methods,
  for example the lasso estimator, post-lasso, \(l_2\)-boosting, or
  forward selection, among others.
\item
  The assumption of density of \(\eta_0\) with respect to some
  dictionary calls for density-based estimators such as the ridge. Mixed
  structures based on sparsity and density suggest the use of elastic
  net or lava.
\item
  If \(\eta_0\) can be well approximated by tree-based methods,
  regression trees and random forests are suitable.
\item
  If \(\eta_0\) can be well approximated by sparse, shallow or deep
  neural networks, \(l_1\)-penalized neural networks, shallow neural
  networks or deep neural networks are attractive.
\end{enumerate}
For most of these ML methods, performance guarantees are available that
make it possible to satisfy the theoretical requirements. For deep
learning results can be found in \citet{farrell2021deep}, for lasso in
\citet{buhlmann2011statistics}. Moreover, if \(\eta_0\) can be well
approximated by at least one model mentioned in the list above, ensemble
or aggregated methods \citep{Wolpert1992, Breiman1996} can be used.
Ensemble and aggregation methods ensure that the performance guarantee
is approximately no worse than the performance of the best method
\citep{van2007super, dudoit2005asymptotics}.

\subsubsection{Third key input: Sample splitting}
The third key input is to use a form of sample splitting at
the stage of producing the estimator of the main parameter \(\theta_0\),
which allows to avoid biases arising from overfitting.

\begin{figure}[t!]
  \centering
\includegraphics[width=0.8\textwidth, trim=15 20 0 0, clip]{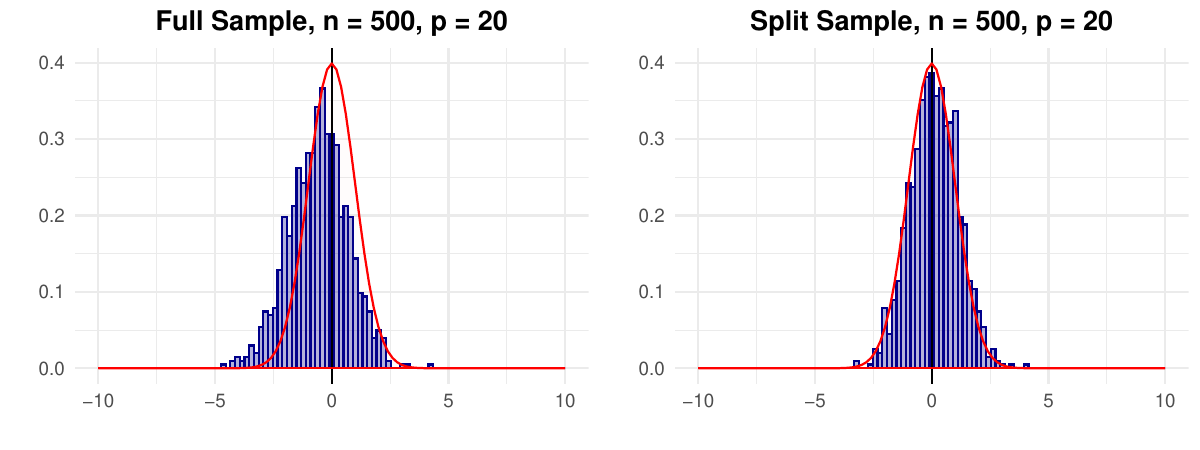}
\caption{%
  Performance of orthogonal estimators based on full sample and sample splitting
in a simulated data example. Left panel: Histogram of the studentized estimator
$\hat{\theta}^{\text{nosplit}}_0$. $\hat{\theta}^{\text{nosplit}}_0$ is based on
estimation of $g_0$ and $m_0$ with random forests and a procedure without
sample-splitting: The entire data set is used for learning the nuisance terms
and estimation of the orthogonal score. Data sets are simulated according to the
data generating process in Section~\ref{anexample}. Data generation and
estimation are repeated 1000 times. Right panel: Histogram of the studentized
DML estimator $\tilde{\theta}_0$. $\tilde{\theta}_0$ is based on estimation of
$g_0$ and $m_0$ with random forests and the cross-fitting described in Algorithm
2. Note that the simulated data sets and parameters of the random forest
learners are identical to those underlying the left panel.
}\label{failure2}
\end{figure}

Biases arising from overfitting could result from using highly complex
fitting methods such as boosting, random forests, ensemble, and hybrid
machine learning methods. We specifically use cross-fitted forms of the
empirical moments, as detailed below in Algorithms 1 and 2, in
estimation of \(\theta_0\). If the same samples would be used to
estimate \(\eta_0\) and the causal parameter \(\theta_0\), we may end up
with very large bias, which we refer to as an overfitting
bias.
While sample splitting is key for the DML approach, other
approaches, like target maximum likelihood, allow for the use of
arbitrary machine learning methods for the estimation of the nuisance
parameters without sample splitting.
The overfitting bias is illustrated in Figure~\ref{failure2}.
The left panel shows the histogram of a studentized
estimator \(\hat{\theta}^{\text{nosplit}}_0\) with \(\hat{\theta}^{\text{nosplit}}_0\)
being obtained from solving the orthogonal score of Equation
\ref{scorepartiallingout} without sample splitting. All observations
are used to learn functions \(g_0\) and \(m_0\) in the PLR model and to
solve the score
\(\frac{1}{N}\sum_i^{N} \psi(W_i; \hat\theta^{\text{nosplit}}_0, \hat{\eta}_0)\).
Consequently, this overfitting bias leads to a considerable shift of
the empirical distribution to the left. The double machine learning
estimator underlying the histogram in the right panel is obtained with
cross-fitting according to Algorithm 2. The sample-splitting procedure
makes it possible to completely eliminate the bias induced by
overfitting.

\section{The double machine learning inference method} \label{dmlinference}

\subsection{Double machine learning for estimation of a causal parameter}

We assume that we have a sample \((W_i)^N_{i_1}\), modeled as i.i.d.
copies of \(W=(Y,D,Z,X)\), whose law is determined by the probability
measure \(P\). We assume that \(N\) is divisible by \(K\) in order to
simplify the notation. Let \(\E_N\) denote the empirical
expectation \begin{align*}
\E_N[g(W)] := \frac{1}{N} \sum_{i=1}^{N}g(W_i).
\end{align*}

\subsubsection{Algorithm 1: DML1 (generic double machine learning with cross-fitting)}

\begin{enumerate}
\item[(1)] Inputs: Choose a model (PLR, PLIV, IRM, IIVM), provide data $(W_i)^N_{i=1}$, a Neyman-orthogonal score function $\psi(W;\theta, \eta)$, which depends on the model being estimated, and specify machine learning methods for $\eta$. 

\item[(2)] Train ML predictors on folds: Take a $K$-fold random partition $(I_k)_{k=1}^{K}$ of observation indices $[N]=\{1, \ldots, N\}$ such that the size of each fold $I_k$ is $n=N/K$. For each $k\in[K]=\{1, \ldots, K\}$, construct a high-quality machine learning estimator 
\begin{align*}
\hat{\eta}_{0,k} = \hat{\eta}_{0,k}\big((W_i)_{i\not\in I_k}\big)
\end{align*}
of $\eta_0$, where $x \mapsto \hat{\eta}_{0,k}(x)$ depends only on the subset of data $(W_i)_{i\not\in I_k}$.

\item[(3)] For each $k\in[K]$, construct the estimator $\check{\theta}_{0,k}$ as the solution to the equation 
\begin{align*}
\frac{1}{n} \sum_{i \in I_k} \psi(W_i; \check{\theta}_{0,k}, \hat{\eta}_{0,k}) = 0.
\end{align*}
The estimate of the causal parameter is obtained via aggregation
\begin{align*}
\tilde{\theta}_0 = \frac{1}{K} \sum_{k=1}^{K} \check{\theta}_{0,k}.
\end{align*}

\item[(4)] Output: The estimate of the causal parameter $\tilde{\theta}_0$ as well as the values of the evaluated score function are returned. 
\end{enumerate}

\subsubsection{Algorithm 2: DML2 (generic double machine learning with cross-fitting)}

\begin{enumerate}
\item[(1)] Inputs: Choose a model (PLR, PLIV, IRM, IIVM), provide data $(W_i)^N_{i=1}$, a Neyman-orthogonal score function $\psi(W;\theta, \eta)$, which depends on the model being estimated, and specify machine learning methods for $\eta$. 

\item[(2)] Train ML predictors on folds: Take a $K$-fold random partition $(I_k)_{k=1}^{K}$ of observation indices $[N]=\{1, \ldots, N\}$ such that the size of each fold $I_k$ is $n=N/K$. For each $k\in[K]=\{1, \ldots, K\}$, construct a high-quality machine learning estimator 
\begin{align*}
\hat{\eta}_{0,k} = \hat{\eta}_{0,k}\big((W_i)_{i\not\in I_k}\big)
\end{align*}
of $\eta_0$, where $x \mapsto \hat{\eta}_{0,k}(x)$ depends only on the subset of data $(W_i)_{i\not\in I_k}$.

\item[(3)] Construct the estimator for the causal parameter $\tilde{\theta}_{0}$ as the solution to the equation
\begin{align*}
\frac{1}{N} \sum_{k=1}^{K} \sum_{i \in I_k} \psi(W_i; \tilde{\theta}_0, \hat{\eta}_{0,k}) = 0.
\end{align*}

\item[(4)] Output: The estimate of the causal parameter $\tilde{\theta}_0$ as well as the values of the evaluated score function are returned. 
\end{enumerate}

Both Algorithm 1 and 2 use out-of-sample predictions generated by ML
learners in order to solve an orthogonal moment condition and, hence,
share the same steps (1) and (2). However, the algorithms differ in the
way the nuisance predictions are plugged into the score function and in
the subsequent solution for \(\theta_0\). In Algorithm 1, the score is
solved on each of the \(K\) folds and the estimate \(\tilde{\theta}_0\)
is obtained by averaging the \(K\) preliminary estimators,
\(\check{\theta}_{0,k}\) with \(k=1, \ldots, K\). According to Algorithm
2, the out-of-sample predictions \(\hat{\eta}_0\) are all plugged into
one score function, which is then solved to obtain the estimate
\(\tilde{\theta}_0\).

\subsubsection{Remark 1: Linear scores}

The score for the models PLR, PLIV, IRM and IIVM are linear in \(\theta\), having the form
\begin{align*}
\psi(W;\theta, \eta) = \psi_a(W; \eta) \theta + \psi_b(W; \eta),
\end{align*} hence the estimator \(\tilde{\theta}_{0,k}\) for DML2
(\(\check{\theta}_{0,k}\) for DML1) takes the form
\begin{align*}
\tilde{\theta}_0 = - \left(\E_N[\psi_a(W; \eta)]\right)^{-1}\E_N[\psi_b(W; \eta)].
\end{align*}
The linear score function representations of the PLR, PLIV, IRM and IIVM
are

PLR with \code{score\ =\ "partialling\ out"}
\begin{align*}
  \begin{aligned}
    \psi_a(W; \eta) &=  - (D - m(X)) (D - m(X)),\\
    \psi_b(W; \eta) &= (Y - \ell(X)) (D - m(X)).
  \end{aligned}
\end{align*}
PLR with \code{score\ =\ "IV-type"}
\begin{align*}
  \begin{aligned}
    \psi_a(W; \eta) &=  - D (D - m(X)),\\
    \psi_b(W; \eta) &= (Y - g(X)) (D - m(X)).
  \end{aligned}
\end{align*}
PLIV with \code{score\ =\ "partialling\ out"}
\begin{align*}
  \begin{aligned}
    \psi_a(W; \eta) &=  - (D - r(X)) (Z - m(X)),\\
    \psi_b(W; \eta) &= (Y - \ell(X)) (Z - m(X)).
  \end{aligned}
\end{align*}
PLIV with \code{score\ =\ "IV-type"}
\begin{align*}
  \begin{aligned}
    \psi_a(W; \eta) &=  - D (Z - m(X)),\\
    \psi_b(W; \eta) &= (Y - g(X)) (Z - m(X)).
  \end{aligned}
\end{align*}
IRM  with \code{score\ =\ "ATE"}
\begin{align*}
  \begin{aligned}
    \psi_a(W; \eta) &=  - 1,\\
    \psi_b(W; \eta) &= g(1,X) - g(0,X) + \frac{D (Y - g(1,X))}{m(X)} - \frac{(1 - D)(Y - g(0,X))}{1 - m(x)}.
  \end{aligned}
\end{align*}
IRM with \code{score\ =\ "ATTE"}
\begin{align*}
  \begin{aligned}
    \psi_a(W; \theta, \eta)&= -\frac{D}{p} \\
    \psi_b(W; \theta, \eta) &= \frac{D (Y - g(0,X))}{p} - \frac{m(X) (1 - D) (Y - g(0,X))}{p(1 - m(x))}
  \end{aligned}
\end{align*}
IIVM with \code{score\ =\ "LATE"}
\begin{align*}
  \begin{aligned}
    \psi_a(W; \eta) &=  - \bigg(r(1,X) - r(0,X) + \frac{Z (D - r(1,X))}{m(X)} - \frac{(1 - Z)(D - r(0,X))}{1 - m(x)} \bigg),\\
    \psi_b(W; \eta) &= g(1,X) - g(0,X) + \frac{Z (Y - g(1,X))}{m(X)} - \frac{(1 - Z)(Y - g(0,X))}{1 - m(x)}.
  \end{aligned}
\end{align*}

\subsubsection{Remark 2: Sample splitting}

In Step (2) of the Algorithm DML1 and DML2, the estimator \(\hat{\eta}_{0,k}\) can generally be an
ensemble or aggregation of several estimators as long as we only use the
data \((W_i)_{i\not\in I_k}\) outside the \(k\)-th fold to construct the
estimators.

\subsubsection{Remark 3: Recommendation}

We have found that \(K=4\) or \(K=5\) to work better than \(K=2\) in a variety of empirical examples
and in simulations. The default for the option \code{n\_folds} that
implements the value of \(K\) is \code{n\_folds=5}. Moreover, we
generally recommend to repeat the estimation procedure multiple times
and use the estimates and standard errors as aggregated over multiple
repetitions as described in \citet[C30-C31]{dml2018}. This aggregation
will be automatically executed if the number of repetitions
\code{n\_rep} is set to a value larger than 1.

The properties of the estimator are as follows.

\begin{theorem} \label{theorem1}
There exist regularity conditions, such that the estimator $\tilde{\theta}_0$ concentrates in a $1/\sqrt{N}$-neighborhood of $\theta_0$ and the sampling error $\sqrt{N}(\tilde{\theta}_0 - \theta_0)$ is approximately normal
\begin{align*}
\sqrt{N}(\tilde{\theta}_0 - \theta_0) \leadsto N(0, \sigma^2),
\end{align*}
with mean zero and variance given by 
\begin{align*}
\begin{aligned}\sigma^2 &= J_0^{-2} \E(\psi^2(W; \theta_0, \eta_0)),\\J_0 &= \E(\psi_a(W; \eta_0)).\end{aligned}
\end{align*}

\end{theorem}

\subsubsection{Algorithm 3: Variance estimation and confidence intervals}

\begin{enumerate}
\item[(1)] Inputs: Use the inputs and outputs from Algorithm 1 (DML1) or Algorithm 2 (DML2). 

\item[(2)] Variance and confidence intervals: Estimate the asymptotic variance of $\tilde{\theta}_0$ by 
\begin{align*}
\begin{aligned}\hat{\sigma}^2 &= \hat{J}_0^{-2} \frac{1}{N} \sum_{k=1}^{K} \sum_{i \in I_k} \big[\psi(W_i; \tilde{\theta}_0, \hat{\eta}_{0,k})\big]^2,\\\hat{J}_0 &= \frac{1}{N} \sum_{k=1}^{K} \sum_{i \in I_k} \psi_a(W_i; \hat{\eta}_{0,k})\end{aligned}
\end{align*}
and form an approximate $(1-\alpha)$ confidence interval, which is asymptotically valid, as 
\begin{align*}
[\tilde{\theta}_0 \pm \Phi^{-1}(1 - \alpha/2) \hat{\sigma} / \sqrt{N}].
\end{align*}
\item[(3)] Output: Output variance estimator and the confidence interval. 
\end{enumerate}

\begin{theorem}
Under the same regularity condition, this interval contains $\theta_0$ for approximately $(1-\alpha)\times 100$ percent of data realizations
\begin{align*}
\Prob\left(\theta_0 \in \left[ \tilde{\theta}_0  \pm \Phi^{-1} (1-\alpha/2) \hat{\sigma}/\sqrt{N} \right] \right) \rightarrow (1-\alpha).
\end{align*}
\end{theorem}

\subsubsection{Remark 4: Brief literature overview on double machine learning}

The presented double machine learning method was developed in
\citet{dml2018}. The idea of using property~\ref{neyman} to construct estimators
and inference procedures that are robust to small mistakes in nuisance
parameters can be traced back to \citet{neyman} and has been used
explicitly or implicitly in the literature on debiased sparsity-based
inference
\citep{belloni2011, belloni2014pivotal, javanmard2014hypothesis, van2014asymptotically, zhang2014, chernozhukov2015valid}
as well as (implicitly) in the classical semi-parametric learning theory
with low-dimensional \(X\)
\citep{levit1975, hasminskii1978, bickel1993efficient, newey1994asymptotic, van2000asymptotic, van2011targeted}.
These references also explain that if we use scores \(\psi\) that are
not Neyman-orthogonal in high dimensional settings, then the resulting
estimators of \(\theta_0\) are not \(1/\sqrt{N}\) consistent and are
generally heavily biased.

\subsubsection{Remark 5: Literature on sample splitting}

Sample
splitting has been used in the traditional semiparametric estimation
literature to establish good properties of semiparametric estimators
under weak conditions
\citep{klaassen1987, schick1986, van2000asymptotic, zheng2011cross}. In
sparse learning problems with high-dimensional \(X\), sample splitting
was employed in \citet{belloni2012sparse}. There and here, the use of
sample splitting results in weak conditions on the estimators of
nuisance parameters, translating into weak assumptions on sparsity in
the case of sparsity-based learning.

\subsubsection{Remark 6: Debiased machine learning}

The presented approach builds upon and generalizes the approach of
\citet{belloni2011}, \citet{zhang2014}, \citet{javanmard2014hypothesis},
\citet{javanmard2014hypothesis}, \citet{javanmard2018debiasing},
\citet{restud}, \citet{bck2014}, \citet{buhlmann2015high}, which
considered estimation of the special case~(\ref{plr1})--(\ref{plr2})
using lasso without cross-fitting. This generalization, by relying upon
cross-fitting, opens up the use of a much broader collection of machine
learning methods and, in the case the lasso is used to estimate the
nuisance functions, allows relaxation of sparsity conditions. All of
these approaches can be seen as ``debiasing'' the estimation of the main
parameter by constructing, implicitly or explicitly, score functions
that satisfy the exact or approximate Neyman orthogonality.

\subsection{Methods for simultaneous inference} \label{simultaneousinf}

In addition to estimation of target causal parameters, standard errors,
and confidence intervals, the package \pkg{DoubleML} provides methods
to perform valid simultaneous inference based on a multiplier bootstrap
procedure introduced in \citet{cck2013} and \citet{cck2014} and
suggested in high-dimensional linear regression models in
\citet{bck2014}. Accordingly, it is possible to (i)~construct simultaneous confidence bands for a potentially large number of causal
parameters and (ii)~adjust $p$~values in a test of multiple hypotheses
based on the inferential procedure introduced above.

We consider a causal PLR with \(p_1\) causal parameters of interest
\(\theta_{0,1}, \ldots, \theta_{0,p_1}\) associated with the treatment
variables \(D_1, \ldots, D_{p_1}\). The parameter of interest
\(\theta_{0,j}\) with \(j=1, \ldots, p_1\) solves a corresponding moment
condition
\begin{align*}
\E\left[\psi_j(W;\theta_{0,j}, \eta_{0,j}) \right]=0,
\end{align*}
as for example considered in \citet{zestim}. To perform
inference in a setting with multiple target coefficients
\(\theta_{0,j}\), the double machine learning procedure implemented in
\pkg{DoubleML} iterates over the target variables of interest. During
estimation of the effect of treatment \(D_j\) on \(Y\) as measured by
the coefficient \(\theta_{0,j}\), the remaining treatment variables
enter the nuisance terms by default, i.e.,~they are added to the set of
control variables \(X\).

\pagebreak

\subsubsection{Algorithm 4: Multiplier bootstrap}

\begin{enumerate}
\item[(1)] Inputs: Use the inputs and outputs from Algorithm 1 (DML1) or Algorithm 2 (DML2) and  Algorithm 3 (Variance estimation) resulting in estimates $\tilde{\theta}_{0,1}, \ldots, \tilde{\theta}_{0,p_1}$, and standard errors $\hat{\sigma}_1, \ldots \hat{\sigma}_{p_1}$.

\item[(2)] Multiplier bootstrap: Generate random weights $\xi_{i}^b$ for each bootstrap repetition $b=1, \ldots, B$ according to a normal (Gaussian) bootstrap, wild bootstrap or exponential bootstrap. Based on the estimated standard errors given by $\hat{\sigma}_j$ and $\hat{J}_{0,j} = \E_N(\psi_{a,j}(W; \eta_{0,j}))$, we obtain bootstrapped versions of the $t$~statistics $t^{*,b}_j$ for $j=1, \ldots, p_1$
\begin{align*}
\begin{aligned}
t^{*,b}_{j} &= \frac{1}{\sqrt{N} \hat{J}_{0,j} \hat{\sigma}_{j}} \sum_{k=1}^{K} \sum_{i \in I_k} \xi_{i}^b  \cdot \psi_j(W_i; \tilde{\theta}_{0,j}, \hat{\eta}_{0,j;k}).\end{aligned}
\end{align*}

\item[(3)] Output:  Output the bootstrapped test statistics. 
\end{enumerate}

\subsubsection{Remark 7: Computational efficiency}

The multiplier bootstrap procedure of \citet{cck2013} and \citet{cck2014} is
computationally efficient because it does not require resampling and
reestimation of the causal parameters. Instead, it is sufficient to
introduce a random pertubation of the score \(\psi\) and solve for
\(\theta_0\), accordingly.

To construct simultaneous \((1-\alpha)\)-confidence bands, the
multiplier bootstrap presented in Algorithm 4 can be used to obtain a
constant \(c_{1-\alpha}\) that will guarantee asymptotic \((1-\alpha\))
coverage
\begin{align} \label{confband}
\left[\tilde\theta_{0,j} \pm c_{1-\alpha} \cdot \hat\sigma_j/\sqrt{N} \right].
\end{align}
The constant \(c_{1-\alpha}\) is obtained in two steps.
\begin{enumerate}
\def\labelenumi{\arabic{enumi}.}
\tightlist
\item
  Calculate the maximum of the absolute values of the bootstrapped
  $t$~statistics, \(t^{*,b}_j\), in every repetition \(b\) with
  \(b=1,\ldots,B\).
\item
  Use the \((1-\alpha)\)-quantile of the \(B\) maxima statistics from
  Step 1 as \(c_{1-\alpha}\) and construct simultaneous confidence bands
  according to Equation~\ref{confband}.
\end{enumerate}
Moreover, it is possible to derive an adjustment method for $p$~values
obtained from a test of multiple hypotheses, including classical
adjustments such as the Bonferroni correction as well as the Romano-Wolf
stepdown procedure \citep{rw2005, rw22005}. The latter is implemented
according to the algorithm for adjustment of $p$~values as provided in
\citet{rw2016} and adapted to high-dimensional linear regression based
on the lasso in \citet{bach2018valid}.

\section{Implementation details} \label{implementationdetails}

In this section, we briefly provide information on the implementation
details such as the class structure, the data backend and the use of
machine learning methods. Section~\ref{illustration} provides a
demonstration of \pkg{DoubleML} in real-data and simulation examples.
More information on the implementation can be found in the DoubleML User
Guide, that is available online at \url{https://docs.doubleml.org/stable/}.
All class methods are documented in the documentation of the
corresponding class, which can be browsed online at \url{https://docs.doubleml.org/r/stable/}
or, for example, by using the commands \code{help(DoubleML)},
\code{help(DoubleMLPLR)}, or \code{help(DoubleMLData)} in \proglang{R}.
For an introduction to \pkg{R6} we refer to the introduction of the online
book for \pkg{mlr3}, available at \url{https://mlr3book.mlr-org.com/intro.html}.

\begin{figure}[t!]
  \centering
\includegraphics[width=0.95\textwidth, trim=10 20 0 20, clip]{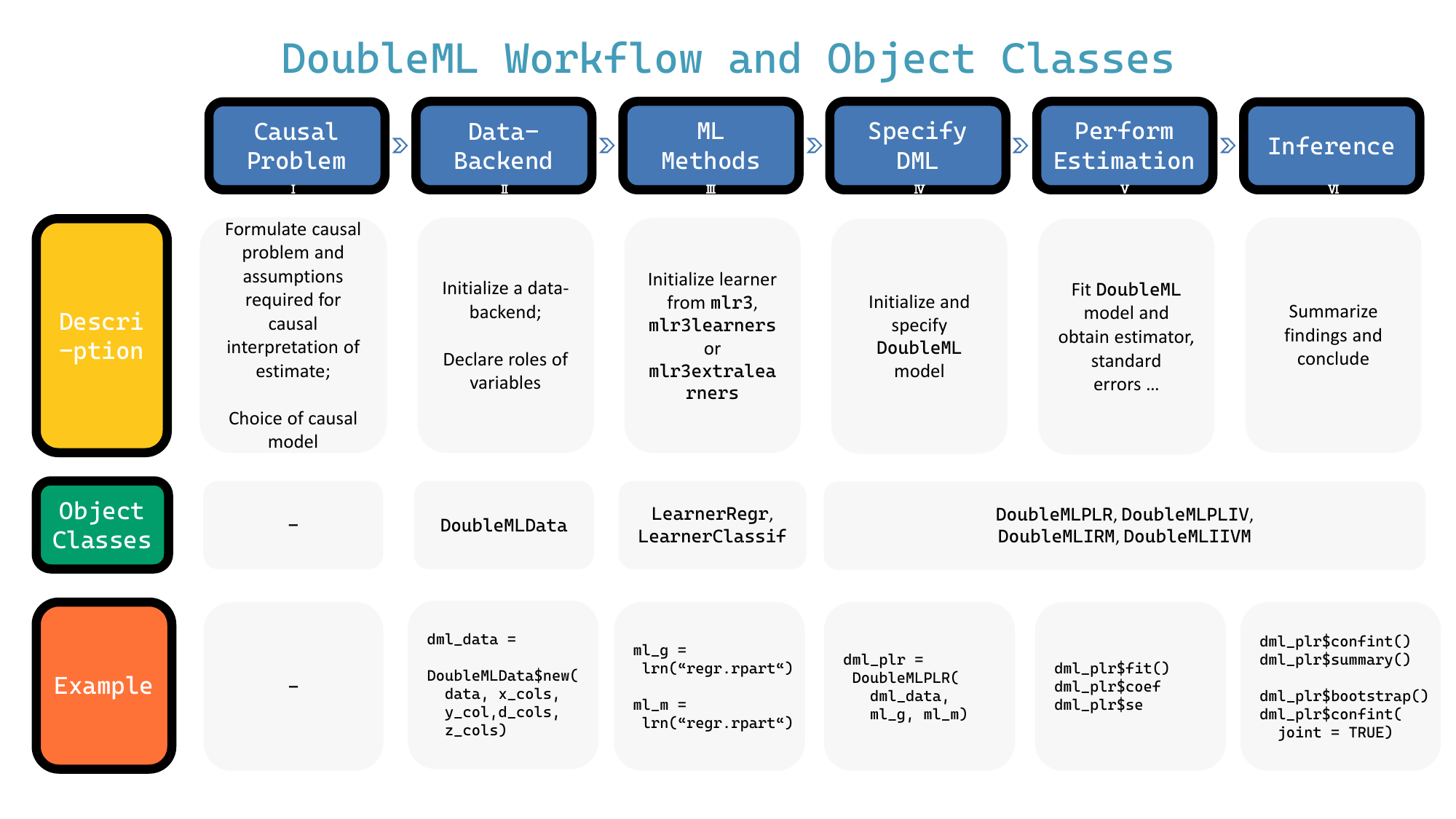}
\caption{%
  Flowchart with main steps and object classes in \pkg{DoubleML}. The flowchart
illustrates the basic steps for estimation of causal parameters with
\pkg{DoubleML}. The diagram contains a short description of the main steps and
lists the object classes used in each step. A short example demonstrates the use
of the object classes and methods.
}\label{workflow}
\end{figure}

\subsection{Object orientation and class structure} \label{classstructure}

\begin{figure}[t!]
\centering
\includegraphics[width=\textwidth]{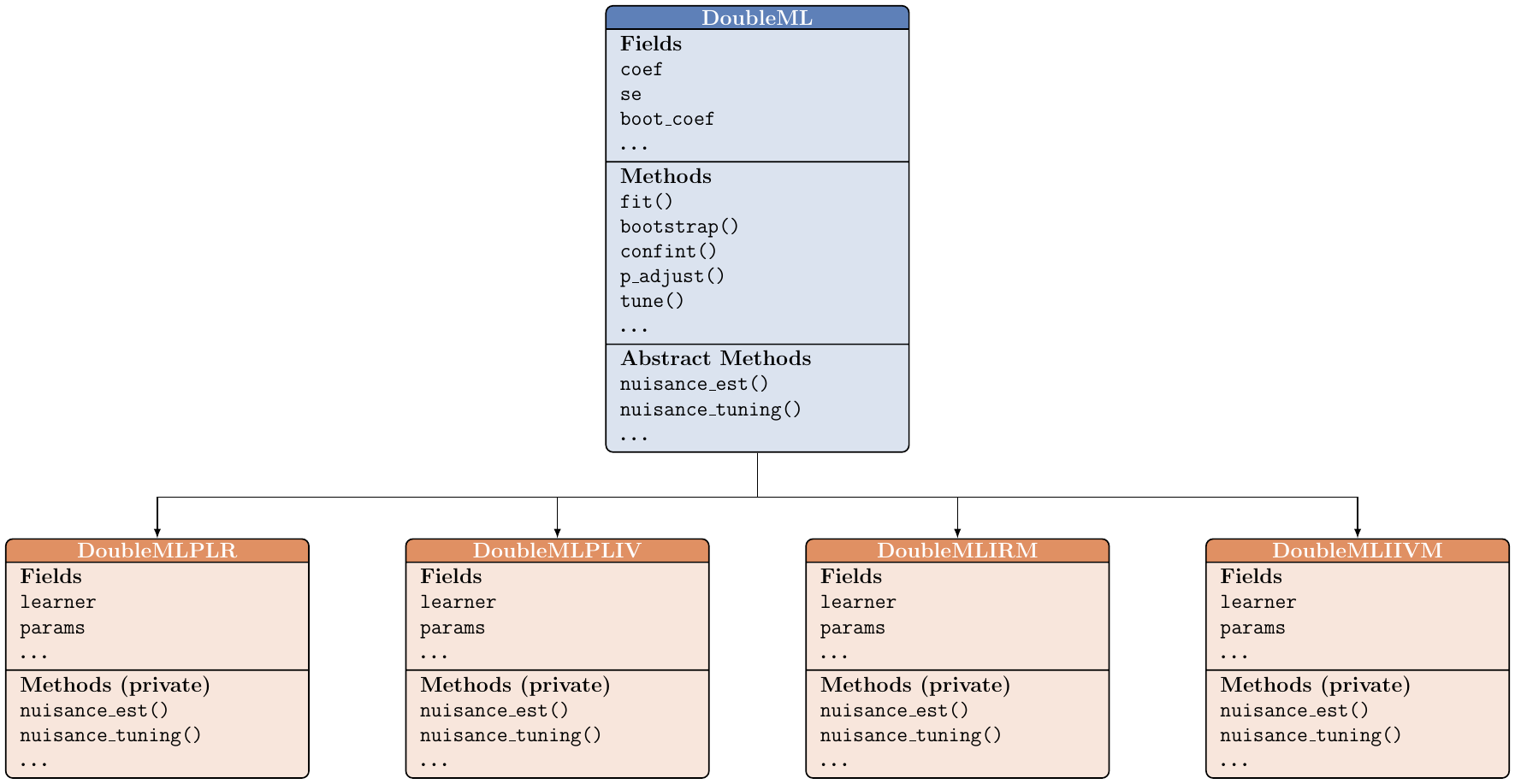}
\caption{Class structure of the \pkg{DoubleML}~package for \proglang{R}.}
\label{masterplan}
\end{figure}

As pointed out in the previous sections, the double machine learning framework provides a general inferential framework in that it covers a plethora of causal models that can be characterized in terms of a Neyman-orthogonal score function $\psi$. In order to design an implementation that is similarly general, the implementation of \pkg{DoubleML} for \proglang{R} is based on object
orientation as enabled by the the \pkg{R6} package \citep{R6}.  The choice of the object orientation provided by \pkg{R6} as compared to alternative approaches (e.g.,~\proglang{S}3 or \proglang{S}4 classes) has been motivated by mainly three reasons: First, we would like to obtain an optimal compatibility with the \pkg{mlr3} ecosystem that is built with \pkg{R6} classes as well. Second, \pkg{R6} makes it possible to use encapsulation, inheritance, active bindings and to distinguish between private and public methods which are important features required in our implementation. Third, the object-oriented implementation of \pkg{DoubleML} makes it possible to achieve a high degree of comparability with its \proglang{Python} twin, which will likely facilitate and accelerate the continuous development of both packages in the future. For an introduction to object orientation in \proglang{R} and the \pkg{R6}~package, we
refer to the vignettes of the \pkg{R6}~package that are available
online at \url{https://r6.r-lib.org/articles/}, Chapter~2.1 of
\citet{mlr3book}, and the chapters on object orientation in
\citet{advr}. The structure of the classes are presented in Figure
\ref{masterplan}. Moreover, the flowchart in Figure~\ref{workflow}
illustrates the main steps of an analysis in \pkg{DoubleML} and links
them to the provided object classes. Figure~\ref{workflow} provides a
short code demonstration, too. The abstract class `\code{DoubleML}'
provides all methods for estimation and inference, for example the
methods \code{fit()}, \code{bootstrap()}, \code{confint()}. All
key components associated with estimation and inference are implemented
in `\code{DoubleML}', for example the sample splitting, the
implementation of Algorithm 1 (DML1) and Algorithm 2 (DML2), the
estimation of the causal parameters, and the computation of the scores
\(\psi(W;\theta, \eta)\). Only the model-specific properties and methods
are allocated at the classes `\code{DoubleMLPLR}' (implementing the
PLR), `\code{DoubleMLPLIV}' (PLIV), `\code{DoubleMLIRM}' (IRM), and
`\code{DoubleMLIIVM}' (IIVM). For example, each of the models has one or
several Neyman-orthogonal score functions that are implemented for the
specific child classes.

\subsection{Data backend and causal model} \label{backend}

The `\code{DoubleMLData}'~class serves as the data backend and
implements the causal model of interest. The user is required to specify
the roles of the variables in a data set at hand. Depending on the
causal model considered, it is necessary to declare the dependent
variable, the treatment variable(s), confounding variables(s), and, in
the case of instrumental variable regression, one or multiple
instruments. The data backend can be initialized from a
\code{data.table} \citep{datatable}. \pkg{DoubleML} provides
wrappers to initialize from `\code{data.frame}' and `\code{matrix}'
objects, as well.

\subsection{Learners, parameters and tuning}

Generally, all learners provided by the packages \pkg{mlr3},
\pkg{mlr3learners} and \pkg{mlr3extralearners} can be used for
estimation of the nuisance functions of the structural models presented
above. An interactive list of supported learners is available at the
\pkg{mlr3extralearners}
website
(\url{https://mlr3extralearners.mlr-org.com/articles/learners/list_learners.html}).
The\linebreak \pkg{mlr3extralearners} package makes it possible to add new
learners, as well. The performance of the double machine learning
estimator \(\tilde\theta_0\) will depend on the predictive quality of
the used machine learning method. Machine learning methods usually have
several (hyper-)parameter that need to be adapted to the specific
application. Tuning of model parameters can be either performed
externally or internally. The latter is implemented in the method
\code{tune()} and is further illustrated in an example in Section
\ref{learners}. Both cases build on the functionalities provided
by the package \pkg{mlr3tuning}.

\subsection{Modifications and extensions}

The flexible architecture of the \pkg{DoubleML} package allows users
to modify the estimation procedure in many regards. Among others, users
can provide customized sample splitting rules after initialization of
the causal model via the method \code{set\_sample\_splitting()}. Moreover, it is possible to adjust the
Neyman-orthogonal score function by externally providing a customized
function via the \code{score} option during initialization of the
causal model object. Short examples for both of these potential extensions are presented in Section~\ref{specific}.

\section{Estimation in real-data and simulated examples} \label{illustration}

In this section, we will first demonstrate the use of \pkg{DoubleML}
in a real-data example, which is based on data from the Pennsylvania
Reemployment Bonus experiment \citep{bilias2000}. This empirical example
has been used in \citet{dml2018}, as well. The goal in the empirical
example is to estimate the causal parameter in a partially linear and an
interactive regression model. We further provide a short example on how
valid simultaneous inference can be performed with \pkg{DoubleML}.
Finally, we present results from a short simulation study as a brief
assessment of the finite-sample performance of the implemented
estimators. Here we want to stress that in real world applications
modelling choices of the estimation of the nuisance parameters and
proper tuning of the parameters are very important. We would like to
mention that the presented examples are mainly included for the purpose
of illustration. In practice, we recommend to carefully choose and tune
the ML learners in terms of their hyperparamaters.

\subsection{Initialization of the data backend}

We begin our real-data example by downloading the Pennsylvania
Reemployment Bonus data set. To do so, we use the call \code{fetch\_bonus()} (a connection to
the internet is required).
\begin{CodeChunk}
\begin{CodeInput}
R> library("DoubleML")
\end{CodeInput}
\end{CodeChunk}
Load data as \code{data.table}.
\begin{CodeChunk}
\begin{CodeInput}
R> dt_bonus <- fetch_bonus(return_type = "data.table")
\end{CodeInput}
\end{CodeChunk}
The output is suppressed for the sake of brevity.
\begin{CodeChunk}
\begin{CodeInput}
R> dt_bonus 
\end{CodeInput}
\end{CodeChunk}

The data backend `\code{DoubleMLData}' can be initialized from a
`\code{data.table}' object by specifying the dependent variable \(Y\)
via a character in \code{y\_col}, the treatment variable(s) \(D\) in
\code{d\_cols}, and the confounders \(X\) via \code{x\_cols}.
Moreover, in IV models, an instrument can be specified via
\code{z\_cols}. In the next step, we assign the roles to the variables
in the data set:
\code{y\_col\ =\ \textquotesingle{}inuidur1\textquotesingle{}} serves
as outcome variable \(Y\), the column
\code{d\_cols\ =\ \textquotesingle{}tg\textquotesingle{}} serves as
treatment variable \(D\) and the columns \code{x\_cols} specify the
confounders.
\begin{CodeChunk}
\begin{CodeInput}
R> obj_dml_data_bonus <- DoubleMLData$new(dt_bonus,
+    y_col = "inuidur1",
+    d_cols = "tg",
+    x_cols = c("female", "black", "othrace", "dep1", "dep2", "q2", "q3",
+    "q4", "q5", "q6", "agelt35", "agegt54", "durable", "lusd", "husd"))
\end{CodeInput}
\end{CodeChunk}
The data backend can be printed to obtain a summary of the main attributes of the\linebreak `\code{DoubleMLData}'~object.
\begin{CodeChunk}
\begin{CodeInput}
R> obj_dml_data_bonus
\end{CodeInput}
\begin{CodeOutput}
================= DoubleMLData Object ==================


------------------ Data summary      ------------------
Outcome variable: inuidur1
Treatment variable(s): tg
Covariates: female, black, othrace, dep1, dep2, q2, q3, q4, q5, q6, agelt35,
  agegt54, durable, lusd, husd
Instrument(s): 
No. Observations: 5099
\end{CodeOutput}
\end{CodeChunk}

Print the data set (output suppressed).
\begin{CodeChunk}
\begin{CodeInput}
R> obj_dml_data_bonus$data
\end{CodeInput}
\end{CodeChunk}

\subsubsection{Remark 8: Wrappers for the data backend}

To initialize an instance of the class `\code{DoubleMLData}' from a `\code{data.frame}' or a collection of
`\code{matrix}'~objects, \pkg{DoubleML} provides the convenient
wrappers \code{double\_ml\_data\_from\_}- \code{data\_frame()} and
\code{double\_ml\_data\_from\_matrix()}. Although the data backend does not provide a formula interface, `\code{DoubleMLData}'~objects can be initialized from a `\code{model.matrix}'
object. The following example demonstrates how users may proceed to specify the causal model by using a `\code{formula}'. We load the Pennsylvania Reemployment Bonus data set as a `\code{data.frame}' and replicate a flexible model specification used in the empirical analysis of \citet[C38--C40]{dml2018}. To flexibly model the nuisance function, we generate all two-way interactions of the control variables.

Load the data as a `\code{data.frame}'.
\begin{CodeChunk}
\begin{CodeInput}
R> df_bonus <- fetch_bonus(return_type = "data.frame")
\end{CodeInput}
\end{CodeChunk}
Print the names of the variables.
\begin{CodeChunk}
\begin{CodeInput}
R> names(df_bonus)
\end{CodeInput}
\begin{CodeOutput}
  [1] "inuidur1" "female"  "black" "othrace" "dep1" "dep2"
  [7] "q2"       "q3"      "q4"    "q5"      "q6"   "agelt35"
 [13] "agegt54"  "durable" "lusd"  "husd"    "tg"
\end{CodeOutput}
\end{CodeChunk}
Specify a `\code{formula}'~object to generate all two-way interactions of the control variables.
\begin{CodeChunk}
\begin{CodeInput}
R> f_flex <- formula(" ~ -1 + (female + black + othrace + dep1 + q2 + q3 +
+    q4 + q5 + q6 + agelt35 + agegt54 + durable + lusd + husd)^2")
\end{CodeInput}
\end{CodeChunk}
Create a `\code{model.matrix}' based on the `\code{formula}'~object.
\begin{CodeChunk}
\begin{CodeInput}
R> X_flex <- model.matrix(f_flex, data = df_bonus)
\end{CodeInput}
\end{CodeChunk}
Initialize using the wrapper \code{double\_ml\_data\_from\_data\_frame()}.
\begin{CodeChunk}
\begin{CodeInput}
R> df_bonus_flex <- data.frame("inuidur1" = df_bonus$inuidur1, X_flex,
+    "tg" = df_bonus$tg)
R> obj_dml_data_bonus_flex <- double_ml_data_from_data_frame(df_bonus_flex,
+    y_col = "inuidur1", d_cols = "tg")
\end{CodeInput}
\end{CodeChunk}
Alternatively, initialize via the wrapper \code{double\_ml\_data\_from\_matrix()}.
\begin{CodeChunk}
\begin{CodeInput}
R> obj_dml_data_bonus_flex2 <- double_ml_data_from_matrix(X = X_flex,
+    y = df_bonus$inuidur1, d = df_bonus$tg)
\end{CodeInput}
\end{CodeChunk}

\subsection{Initialization of the causal model} \label{initialization}

To initialize a PLR model, we have to provide a learner for each
nuisance part in the model in Equation~\ref{plr1}--\ref{plr2}. In \proglang{R},
this is done by providing learners to the arguments \code{ml\_m} for
nuisance part \(m\) and \code{ml\_l} for nuisance part \(\ell\). We
can pass a learner as instantiated in \pkg{mlr3} and
\pkg{mlr3learners}, for example a random forest as provided by the \proglang{R}
package \pkg{ranger} \citep{ranger}. Previous installation of
\pkg{ranger} is required. Moreover, we can specify the score (allowed
choices for PLR are \code{"partialling\ out"} or \code{"IV-type"})
and the algorithm via the option \code{dml\_procedure} (allowed
choices \code{"dml1"} and \code{"dml2"}) . Optionally, it is
possible to change the number of folds used for sample splitting through
\code{n\_folds} and the number of repetitions via \code{n\_rep}, if
the sample splitting and estimation procedure should be repeated.

Set a seed for replication of the sample split.
\begin{CodeChunk}
\begin{CodeInput}
R> set.seed(31415)
R> learner_l <- lrn("regr.ranger", num.trees = 500, min.node.size = 2,
+    max.depth = 5)
R> learner_m <- lrn("regr.ranger", num.trees = 500, min.node.size = 2,
+    max.depth = 5)
R> doubleml_bonus <- DoubleMLPLR$new(obj_dml_data_bonus, ml_l = learner_l,
+    ml_m = learner_m, score = "partialling out", dml_procedure = "dml1",
+    n_folds = 5, n_rep = 1)
R> doubleml_bonus
\end{CodeInput}
\begin{CodeOutput}
================= DoubleMLPLR Object ==================


------------------ Data summary      ------------------
Outcome variable: inuidur1
Treatment variable(s): tg
Covariates: female, black, othrace, dep1, dep2, q2, q3, q4, q5, q6, agelt35,
  agegt54, durable, lusd, husd
Instrument(s): 
No. Observations: 5099

------------------ Score & algorithm ------------------
Score function: partialling out
DML algorithm: dml1

------------------ Machine learner   ------------------
ml_l: regr.ranger
ml_m: regr.ranger

------------------ Resampling        ------------------
No. folds: 5
No. repeated sample splits: 1
Apply cross-fitting: TRUE

------------------ Fit summary       ------------------
 
\end{CodeOutput}
\end{CodeChunk}

\subsection{Estimation of the causal parameter in a PLR model} \label{estimplr}

To perform estimation, call the \code{fit()} method. The output can be
summarized using the method \code{summary()}.
\begin{CodeChunk}
\begin{CodeInput}
R> doubleml_bonus$fit()
R> doubleml_bonus$summary()
\end{CodeInput}
\begin{CodeOutput}
Estimates and significance testing of the effect of target variables
   Estimate. Std. Error t value Pr(>|t|)  
tg  -0.07438    0.03543  -2.099   0.0358 *
---
Signif. codes:  0 '***' 0.001 '**' 0.01 '*' 0.05 '.' 0.1 ' ' 1
\end{CodeOutput}
\end{CodeChunk}
Hence, there is evidence to reject the null hypothesis that
\(\theta_{0, tg}=0\) at the 5\% significance level. The estimated
coefficient and standard errors can be accessed via the attributes
\code{coef} and \code{se} of the object \code{doubleml\_bonus}.
\begin{CodeChunk}
\begin{CodeInput}
R> doubleml_bonus$coef
\end{CodeInput}
\begin{CodeOutput}
         tg 
-0.07438411 
\end{CodeOutput}
\begin{CodeInput}
R> doubleml_bonus$se
\end{CodeInput}
\begin{CodeOutput}
        tg 
0.03543316 
\end{CodeOutput}
\end{CodeChunk}
After completed estimation, we can access the resulting score
\(\psi(W_i; \tilde{\theta}_0, \hat{\eta}_0)\) or the components
\(\psi_a(W_i; \hat{\eta}_0)\) and \(\psi_b(W_i; \hat{\eta}_0)\). The
estimated score for the first 5 observations can be obtained via the public field \code{psi}. \code{psi} is an \code{array} with \code{dim = c(n\_obs, n\_rep, n\_treat)} with
\begin{itemize}
\tightlist
\item \code{n\_obs}: number of observations in the data,
\item \code{n\_rep}: number of repetitions (sample splitting),
\item \code{n\_treat}: number of treatment variables.
\end{itemize}
\begin{CodeChunk}
\begin{CodeInput}
R> doubleml_bonus$psi[1:5, 1, 1] 
\end{CodeInput}
\begin{CodeOutput}
[1] -0.2739454  0.7444154 -0.4509358  0.1813111 -0.3699474
\end{CodeOutput}
\end{CodeChunk}
Similarly, the components of the score \(\psi_a(W_i; \hat{\eta}_0)\) and
\(\psi_b(W_i; \hat{\eta}_0)\) are available as fields.
\begin{CodeChunk}
\begin{CodeInput}
R> doubleml_bonus$psi_a[1:5, 1, 1] 
\end{CodeInput}
\begin{CodeOutput}
[1] -0.0981220 -0.1353987 -0.1276526 -0.4272341 -0.1126174
\end{CodeOutput}
\begin{CodeInput}
R> doubleml_bonus$psi_b[1:5, 1, 1] 
\end{CodeInput}
\begin{CodeOutput}
[1] -0.2812441  0.7343439 -0.4604311  0.1495317 -0.3783243
\end{CodeOutput}
\end{CodeChunk}
To construct a \((1-\alpha)\) confidence interval, we use the
\code{confint()} method.
\begin{CodeChunk}
\begin{CodeInput}
R> doubleml_bonus$confint(level = 0.95)
\end{CodeInput}
\begin{CodeOutput}
        2.5 %       97.5 %
tg -0.1438318 -0.004936395
\end{CodeOutput}
\end{CodeChunk}

\subsection{Estimation of the causal parameter in an IRM model} \label{estimirm}

The treatment variable \(D\) in the Pennsylvania Reemployment Bonus
example is binary. Accordingly, it is possible to estimate an IRM model.
Since the IRM requires estimation of the propensity score
\(\Prob(D\mid X)\), we have to specify a classifier for the nuisance
part \(m_0\).

Initialize a classifier for estimation of the propensity score and create a new instance of a causal model, here an IRM.
\begin{CodeChunk}
\begin{CodeInput}
R> learner_g <- lrn("regr.ranger", num.trees = 500, min.node.size = 2,
+    max.depth = 5)
R> learner_classif_m <- lrn("classif.ranger", num.trees = 500,
+    min.node.size = 2, max.depth = 5)
R> doubleml_irm_bonus <- DoubleMLIRM$new(obj_dml_data_bonus,
+    ml_g = learner_g, ml_m = learner_classif_m, score = "ATE",
+    dml_procedure = "dml1", n_folds = 5, n_rep = 1)
\end{CodeInput}
\end{CodeChunk}
The output is suppressed for the sake of brevity.
\begin{CodeChunk}
\begin{CodeInput}
R> doubleml_irm_bonus
\end{CodeInput}
\end{CodeChunk}
To perform estimation, call the \code{fit()} method. The output can be
summarized using the method \code{summary()}.
\begin{CodeChunk}
\begin{CodeInput}
R> doubleml_irm_bonus$fit()
R> doubleml_irm_bonus$summary()
\end{CodeInput}
\begin{CodeOutput}
Estimates and significance testing of the effect of target variables
   Estimate. Std. Error t value Pr(>|t|)  
tg  -0.07193    0.03554  -2.024    0.043 *
---
Signif. codes:  0 '***' 0.001 '**' 0.01 '*' 0.05 '.' 0.1 ' ' 1
\end{CodeOutput}
\end{CodeChunk}
The estimated coefficient is very similar to the estimate of the PLR
model and our conclusions remain unchanged.

\subsection{Simultaneous inference in a simulated data example} \label{siminf_example}

We consider a simulated example of a PLR model to illustrate the use of
methods for simultaneous inference. First, we will generate a sparse
linear model with only three variables having a non-zero effect on the
dependent variable.
\begin{CodeChunk}
\begin{CodeInput}
R> set.seed(3141)
R> n_obs <- 500
R> n_vars <- 100
R> theta <- rep(3, 3)
\end{CodeInput}
\end{CodeChunk}
Generate a \code{data.frame} and use the corresponding wrapper.
\begin{CodeChunk}
\begin{CodeInput}
R> X <- matrix(stats::rnorm(n_obs * n_vars), nrow = n_obs, ncol = n_vars)
R> y <- X[, 1:3, drop = FALSE] %*% theta  + stats::rnorm(n_obs)
R> df <- data.frame(y, X)
\end{CodeInput}
\end{CodeChunk}
We use the wrapper \code{double\_ml\_data\_from\_data\_frame()} to
specify a data backend that assigns the first 10 columns of \(X\) as
treatment variables and declares the remaining columns as confounders.
\begin{CodeChunk}
\begin{CodeInput}
R> doubleml_data <- double_ml_data_from_data_frame(df, y_col = "y",
+    d_cols = c("X1", "X2", "X3", "X4", "X5", "X6", "X7", "X8", "X9", "X10"))
\end{CodeInput}
\begin{CodeOutput}
Set treatment variable d to X1.
\end{CodeOutput}
\end{CodeChunk}
The output is suppressed for the sake of brevity.
\begin{CodeChunk}
\begin{CodeInput}
R> doubleml_data
\end{CodeInput}
\end{CodeChunk}
A sparse setting suggests the use of the lasso learner. Here, we use the
lasso estimator with cross-validated choice of the penalty parameter
\(\lambda\) as provided in the \pkg{glmnet} package for \proglang{R}
\citep{glmnet}.

Output messages during fitting are suppressed.
\begin{CodeChunk}
\begin{CodeInput}
R> ml_l <- lrn("regr.cv_glmnet", s = "lambda.min")
R> ml_m <- lrn("regr.cv_glmnet", s = "lambda.min")
R> doubleml_plr <- DoubleMLPLR$new(doubleml_data, ml_l, ml_m)
R> doubleml_plr$fit()
R> doubleml_plr$summary()
\end{CodeInput}
\begin{CodeOutput}
Estimates and significance testing of the effect of target variables
    Estimate. Std. Error t value Pr(>|t|)    
X1   3.017802   0.046180  65.348   <2e-16 ***
X2   3.025812   0.042683  70.891   <2e-16 ***
X3   3.000914   0.045849  65.452   <2e-16 ***
X4  -0.034815   0.040955  -0.850   0.3953    
X5   0.035118   0.048132   0.730   0.4656    
X6   0.002171   0.044622   0.049   0.9612    
X7  -0.036129   0.046798  -0.772   0.4401    
X8   0.020361   0.044048   0.462   0.6439    
X9  -0.019439   0.043180  -0.450   0.6526    
X10  0.076180   0.043682   1.744   0.0812 .  
---
Signif. codes:  0 '***' 0.001 '**' 0.01 '*' 0.05 '.' 0.1 ' ' 1
\end{CodeOutput}
\end{CodeChunk}
The multiplier bootstrap procedure can be executed using the
\code{bootstrap()} method where the option \code{method} specifies
the choice of the random pertubations and \code{n\_rep\_boot} the
number of bootstrap repetitions.
\begin{CodeChunk}
\begin{CodeInput}
R> doubleml_plr$bootstrap(method = "normal", n_rep_boot = 1000)
\end{CodeInput}
\end{CodeChunk}
The resulting bootstrapped $t$~statistics are
available via the field \code{boot\_t\_stat}.
To construct a simultaneous confidence interval, we set the option
\code{joint\ =\ TRUE} when calling the \code{confint()} method.
\begin{CodeChunk}
\begin{CodeInput}
R> doubleml_plr$confint(joint = TRUE)
\end{CodeInput}
\begin{CodeOutput}
          2.5 %     97.5 %
X1   2.88766757 3.14793595
X2   2.90553386 3.14609021
X3   2.87171334 3.13011430
X4  -0.15022399 0.08059423
X5  -0.10051468 0.17075155
X6  -0.12357302 0.12791441
X7  -0.16800517 0.09574654
X8  -0.10376590 0.14448792
X9  -0.14111984 0.10224143
X10 -0.04691574 0.19927524
\end{CodeOutput}
\end{CodeChunk}
The correction of the $p$~values of a joint hypotheses test on the
considered causal parameters is implemented in the method
\code{p\_adjust()}. By default, the adjustment procedure specified in
the option \code{method} is the Romano-Wolf stepdown procedure.
\begin{CodeChunk}
\begin{CodeInput}
R> doubleml_plr$p_adjust(method = "romano-wolf")
\end{CodeInput}
\begin{CodeOutput}
       Estimate.  pval
X1   3.017801759 0.000
X2   3.025812035 0.000
X3   3.000913821 0.000
X4  -0.034814877 0.942
X5   0.035118436 0.942
X6   0.002170694 0.961
X7  -0.036129317 0.942
X8   0.020361010 0.951
X9  -0.019439209 0.951
X10  0.076179750 0.451
\end{CodeOutput}
\end{CodeChunk}
Alternatively, the correction methods provided in the \pkg{stats}
function \code{p.adjust} can be applied, for example the Bonferroni,
Bonferroni-Holm, or Benjamini-Hochberg correction. For example a
Bonferroni correction could be performed by specifying
\code{method\ =\ "bonferroni"}.
\begin{CodeChunk}
\begin{CodeInput}
R> doubleml_plr$p_adjust(method = "bonferroni")
\end{CodeInput}
\begin{CodeOutput}
       Estimate.      pval
X1   3.017801759 0.0000000
X2   3.025812035 0.0000000
X3   3.000913821 0.0000000
X4  -0.034814877 1.0000000
X5   0.035118436 1.0000000
X6   0.002170694 1.0000000
X7  -0.036129317 1.0000000
X8   0.020361010 1.0000000
X9  -0.019439209 1.0000000
X10  0.076179750 0.8116808
\end{CodeOutput}
\end{CodeChunk}

\subsection{Learners, parameters and tuning} \label{learners}

The performance of the final double machine learning estimator depends
on the predictive performance of the underlying ML method. First, we
briefly show how externally tuned parameters can be passed to the
learners in \pkg{DoubleML}. Second, it is demonstrated how the
parameter tuning can be done internally by \pkg{DoubleML}.

\subsubsection{External tuning and parameter passing}

Section 3 of the \pkg{mlr3} book \citep{mlr3book} provides a step-by-step
introduction to the powerful tuning functionalities of the
\pkg{mlr3tuning} package. Accordingly, it is possible to manually
reconstruct the \pkg{mlr3} regression and classification problems,
which are internally handled in \pkg{DoubleML}, and to perform
parameter tuning accordingly. One advantage of this procedure is that it
allows users to fully exploit the powerful benchmarking and tuning tools
of \pkg{mlr3} and \pkg{mlr3tuning}.

Consider the sparse regression example from above. We will briefly
consider a setting where we explicitly set the parameter \(\lambda\) for
a \pkg{glmnet} estimator rather than using the interal
cross-validated choice with \code{cv\_glmnet}.

Suppose for simplicity, some external tuning procedure resulted in an
optimal value of \(\lambda=0.1\) for nuisance part \(m\) and
\(\lambda=0.09\) for nuisance part \(\ell\) for the first treatment
variable and \(\lambda=0.095\) and \(\lambda=0.085\) for the second
variable, respectively. After initialization of the model object, we can
set the parameter values using the method
\code{set\_ml\_nuisance\_params()}.
\begin{CodeChunk}
\begin{CodeInput}
R> ml_l <- lrn("regr.glmnet")
R> ml_m <- lrn("regr.glmnet")
R> doubleml_plr <- DoubleMLPLR$new(doubleml_data, ml_l, ml_m)
\end{CodeInput}
\end{CodeChunk}
To set the values, we have to specify the treatment variable and the
nuisance part. If no values are set, the default values are used. Note that variable names are overwritten by the wrapper for the matrix interface.
\begin{CodeChunk}
\begin{CodeInput}
R> doubleml_plr$set_ml_nuisance_params("ml_m", "X1",
+    param = list("lambda" = 0.1))
R> doubleml_plr$set_ml_nuisance_params("ml_l", "X1",
+    param = list("lambda" = 0.09))
R> doubleml_plr$set_ml_nuisance_params("ml_m", "X2",
+    param = list("lambda" = 0.095))
R> doubleml_plr$set_ml_nuisance_params("ml_l", "X2",
+    param = list("lambda" = 0.085))
\end{CodeInput}
\end{CodeChunk}
All externally specified parameters can be retrieved from the field
\code{params}. The output is omitted for the sake of brevity.
\begin{CodeChunk}
\begin{CodeInput}
R> str(doubleml_plr$params)
R> doubleml_plr$fit()
R> doubleml_plr$summary()
\end{CodeInput}
\begin{CodeOutput}
Estimates and significance testing of the effect of target variables
    Estimate. Std. Error t value Pr(>|t|)    
X1   3.041094   0.060030  50.660   <2e-16 ***
X2   2.993916   0.054590  54.844   <2e-16 ***
X3   2.993419   0.055144  54.283   <2e-16 ***
X4  -0.035201   0.040637  -0.866    0.386    
X5   0.021541   0.047569   0.453    0.651    
X6  -0.006652   0.044715  -0.149    0.882    
X7  -0.039650   0.046823  -0.847    0.397    
X8   0.011146   0.044037   0.253    0.800    
X9  -0.021342   0.043237  -0.494    0.622    
X10  0.084426   0.043641   1.935    0.053 .  
---
Signif. codes:  0 '***' 0.001 '**' 0.01 '*' 0.05 '.' 0.1 ' ' 1
\end{CodeOutput}
\end{CodeChunk}

\subsubsection{Internal tuning and parameter passing}

An alternative to external tuning and parameter provisioning is to
perform the tuning internally. The advantage of this approach is that
users do not have to specify the underlying prediction problems
manually. Instead, \pkg{DoubleML} uses the underlying data backend to
ensure that the machine learning methods are tuned for the specific
model under consideration and, hence, to possibly avoid mistakes. We
initialize our structural model object with the learner. At this stage,
we do not specify any parameters.

Load required packages for tuning and set logger to omit messages during tuning and fitting.
\begin{CodeChunk}
\begin{CodeInput}
R> library("paradox")
R> library("mlr3tuning")
R> lgr::get_logger("mlr3")$set_threshold("warn")
R> lgr::get_logger("bbotk")$set_threshold("warn")
R> set.seed(1234)
R> ml_l <- lrn("regr.glmnet")
R> ml_m <- lrn("regr.glmnet")
R> doubleml_plr <- DoubleMLPLR$new(doubleml_data, ml_l, ml_m)
\end{CodeInput}
\end{CodeChunk}
To perform parameter tuning, we provide a grid of values used for
evaluation for each of the nuisance parameters. To set up a grid of
values, we specify a named list with names corresponding to the learner
names of the nuisance part (see method \code{learner\_names()}). The
elements in the list are objects of the class `\code{ParamSet}' of the
\pkg{paradox} package \citep{paradox}.
\begin{CodeChunk}
\begin{CodeInput}
R> par_grids <- list(
  +   "ml_l" = ps(lambda = p_dbl(lower = 0.05, upper = 0.1)),
  +   "ml_m" = ps(lambda = p_dbl(lower = 0.05, upper = 0.1)))
\end{CodeInput}
\end{CodeChunk}
The hyperparameter tuning is performed according to options passed
through a named list \code{tune\_settings}. The entries in the list
specify options during parameter tuning with \pkg{mlr3tuning}:
\begin{itemize}
\item
  \code{terminator} is a \code{bbotk::Terminator}~object passed to
  \pkg{mlr3tuning} that manages the budget to solve the tuning
  problem.
\item
  \code{algorithm} is an object of class `\code{mlr3tuning::Tuner}'
  and specifies the tuning algorithm. Alternatively, algorithm can be a
  \code{character()} that is used as an argument in the wrapper
  \pkg{mlr3tuning} call \code{tnr(algorithm)}. The `\code{Tuner}'
  class in \pkg{mlr3tuning} supports grid search, random search,
  generalized simulated annealing and non-linear optimization.
\item
  \code{rsmp\_tune} is an object of class `\code{mlr3::Resampling}'
  that specifies the resampling method for evaluation, for example
  \code{rsmp("cv",\ folds\ =\ 5)} implements 5-fold cross-validation.
  \code{rsmp("holdout",\ ratio\ =\ 0.8)} implements an evaluation
  based on a hold-out sample that contains 20 percent of the
  observations. By default, 5-fold cross-validation is performed.
\item
  \code{measure} is a named list containing the measures used for
  tuning of the nuisance components. The names of the entries must match
  the learner names (see method \code{learner\_names()}). The entries
  in the list must either be objects of class `\code{mlr3::Measure}' or keys
  passed to \code{msr()}. If \code{measure} is not provided by the
  user, the mean squared error is used for regression models and the
  classification error for binary outcomes, by default.
\end{itemize}
In the next code chunk, the value of the parameter \(\lambda\) is tuned
via grid search in the range 0.05 to 0.1 at a resolution of
11.
The resulting grid has 11 equally spaced values ranging
from a minimum value of 0.05 to a maximum value of 0.1. Type
\code{generate\_design\_grid(par\_grids\$ml\_l,\ resolution\ =\ 11)}
to access the grid for nuisance function \code{ml\_l}.
To evaluate
the predictive performance in both nuisance functions, the
cross-validated mean squared error is used.

Provide tune settings.
\begin{CodeChunk}
\begin{CodeInput}
R> tune_settings <- list(terminator = trm("evals", n_evals = 100),
+    algorithm = tnr("grid_search", resolution = 11),
+    rsmp_tune = rsmp("cv", folds = 5),
+    measure = list("ml_l" = msr("regr.mse"), "ml_m" = msr("regr.mse")))
\end{CodeInput}
\end{CodeChunk}
With these parameters we can run the tuning by calling the \code{tune()}
method for `\code{DoubleML}' objects.

Execution might take around 50 seconds.
\begin{CodeChunk}
\begin{CodeInput}
R> doubleml_plr$tune(param_set = par_grids, tune_settings = tune_settings)
\end{CodeInput}
\end{CodeChunk}
Output omitted for the sake of brevity, available in the appendix. Access tuning results for target variable \code{X1}.
\begin{CodeChunk}
\begin{CodeInput}
R> doubleml_plr$tuning_res$X1
\end{CodeInput}
\end{CodeChunk}
Access tuned parameters (output suppressed).
\begin{CodeChunk}
\begin{CodeInput}
R> str(doubleml_plr$params)
\end{CodeInput}
\end{CodeChunk}
Estimate model and call the \code{summary()} method.
\begin{CodeChunk}
\begin{CodeInput}
R> doubleml_plr$fit()
R> doubleml_plr$summary()
\end{CodeInput}

\vspace*{-0.35cm}

\begin{CodeOutput}
Estimates and significance testing of the effect of target variables
    Estimate. Std. Error t value Pr(>|t|)    
X1   3.028980   0.059701  50.736   <2e-16 ***
X2   3.008650   0.054301  55.407   <2e-16 ***
X3   2.960571   0.053082  55.773   <2e-16 ***
X4  -0.037859   0.040976  -0.924   0.3555    
X5   0.030018   0.047880   0.627   0.5307    
X6   0.003451   0.044419   0.078   0.9381    
X7  -0.025875   0.046936  -0.551   0.5814    
X8   0.022008   0.044172   0.498   0.6183    
X9  -0.014251   0.043765  -0.326   0.7447    
X10  0.088653   0.043691   2.029   0.0424 *  
---
Signif. codes:  0 '***' 0.001 '**' 0.01 '*' 0.05 '.' 0.1 ' ' 1
\end{CodeOutput}
\end{CodeChunk}
By default, the parameter tuning is performed on the whole sample, for
example in the case of \(K_{\text{tune}}\)-fold cross-validation, the entire
sample is split into \(K_{\text{tune}}\) folds for evaluation of the
cross-validated error. Alternatively, each of the \(K\) folds used in
the cross-fitting procedure could be split up into \(K_{\text{tune}}\) subfolds
that are then used for evaluation of the candidate models. As a result,
the choice of the tuned parameters will be fold-specific. To perform
fold-specific tuning, users can set the option
\code{tune\_on\_folds\ =\ TRUE} when calling the method
\code{tune()}.

\subsection{Specifications and modifications of double machine learning} \label{specific}

The flexible architecture of the \pkg{DoubleML} package allows users
to modify the estimation procedure in many regards. We will shortly
present two examples on how users can adjust the double machine learning
framework to their needs in terms of the sample splitting procedure and
the score function.

\subsubsection{Sample splitting}

By default, \pkg{DoubleML} performs cross-fitting as presented in
Algorithms 1 and 2. Alternatively, all implemented models allow a
partition to be provided externally via the method
\code{set\_sample\_splitting()}. Note that by setting
\code{draw\_sample\_splitting\ =\ FALSE} one can prevent that a
partition is drawn during initialization of the model object. The
following calls are equivalent. In the first sample code, we use the
standard interface and draw the sample-splitting with \(K=4\) folds
during initialization of the `\code{DoubleMLPLR}'~object.

First generate some data and initialize ML learners and a data backend.

\vspace*{-0.25cm}

\begin{CodeChunk}
\begin{CodeInput}
R> learner <- lrn("regr.ranger", num.trees = 100, mtry = 20,
+    min.node.size = 2, max.depth = 5)
R> ml_l <- learner
R> ml_m <- learner
R> data <- make_plr_CCDDHNR2018(alpha = 0.5, n_obs = 100,
+    return_type = "data.table")
R> doubleml_data <- DoubleMLData$new(data, y_col = "y", d_cols = "d")
R> set.seed(314)
R> doubleml_plr_internal <- DoubleMLPLR$new(doubleml_data, ml_l, ml_m,
+    n_folds = 4)
R> doubleml_plr_internal$fit()
R> doubleml_plr_internal$summary()
\end{CodeInput}
\begin{CodeOutput}
Estimates and significance testing of the effect of target variables
  Estimate. Std. Error t value Pr(>|t|)    
d    0.4892     0.1024   4.776 1.79e-06 ***
---
Signif. codes:  0 '***' 0.001 '**' 0.01 '*' 0.05 '.' 0.1 ' ' 1
\end{CodeOutput}
\end{CodeChunk}
In the second sample code, we manually specify a sampling scheme using
the `\code{mlr3::Resampling}' class. Alternatively, users can provide a
nested list that has the following structure:
\begin{itemize}
\tightlist
\item
  The length of the outer list must match with the desired number of
  repetitions of the sample-splitting, i.e.,~\code{n\_rep}.
\item
  The inner list is a named list of length 2 specifying the
  \code{test\_ids} and \code{train\_ids}. The named entries
  \code{test\_ids} and \code{train\_ids} are lists of the same
  length,
  \begin{itemize}
  \tightlist
  \item
    \code{train\_ids} is a list of length \code{n\_folds} that
    specifies the indices of the observations used for model fitting in
    each fold,
  \item
    \code{test\_ids} is a list of length \code{n\_folds} that
    specifies the indices of the observations used for calculation of
    the score in each fold.
  \end{itemize}
\end{itemize}
Set up a task and cross-validation resampling scheme in \pkg{mlr3}.
\begin{CodeChunk}
\begin{CodeInput}
R> doubleml_plr_external <- DoubleMLPLR$new(doubleml_data, ml_l, ml_m,
+    draw_sample_splitting = FALSE)
R> set.seed(314)
R> my_task <- Task$new("help task", "regr", data)
R> my_sampling <- rsmp("cv", folds = 4)$instantiate(my_task)
R> train_ids <- lapply(1:4, function(x) my_sampling$train_set(x))
R> test_ids <- lapply(1:4, function(x) my_sampling$test_set(x))
R> smpls = list(list(train_ids = train_ids, test_ids = test_ids))
\end{CodeInput}
\end{CodeChunk}
Structure of the specified sampling scheme.
\begin{CodeChunk}
\begin{CodeInput}
R> str(smpls)
\end{CodeInput}
\begin{CodeOutput}
List of 1
 $ :List of 2
  ..$ train_ids:List of 4
  .. ..$ : int [1:75] 1 7 11 18 19 20 21 31 32 37 ...
  .. ..$ : int [1:75] 10 15 16 22 26 35 38 40 41 46 ...
  .. ..$ : int [1:75] 10 15 16 22 26 35 38 40 41 46 ...
  .. ..$ : int [1:75] 10 15 16 22 26 35 38 40 41 46 ...
  ..$ test_ids :List of 4
  .. ..$ : int [1:25] 10 15 16 22 26 35 38 40 41 46 ...
  .. ..$ : int [1:25] 1 7 11 18 19 20 21 31 32 37 ...
  .. ..$ : int [1:25] 3 5 6 8 17 24 25 28 29 34 ...
  .. ..$ : int [1:25] 2 4 9 12 13 14 23 27 30 33 ...
\end{CodeOutput}
\end{CodeChunk}
Fit the model and summarize.
\begin{CodeChunk}
\begin{CodeInput}
R> doubleml_plr_external$set_sample_splitting(smpls)
R> doubleml_plr_external$fit()
R> doubleml_plr_external$summary()
\end{CodeInput}
\begin{CodeOutput}
Estimates and significance testing of the effect of target variables
  Estimate. Std. Error t value Pr(>|t|)    
d    0.4892     0.1024   4.776 1.79e-06 ***
---
Signif. codes:  0 '***' 0.001 '**' 0.01 '*' 0.05 '.' 0.1 ' ' 1
\end{CodeOutput}
\end{CodeChunk}
Setting the option \code{apply\_cross\_fitting\ =\ FALSE} at the
instantiation of the causal model allows double machine learning being
performed without cross-fitting. It results in randomly splitting the
sample into two parts. The first half of the data is used for the
estimation of the nuisance models with the machine learning methods and
the second half for estimating the causal parameter, i.e.,~solution of
the score. Note that cross-fitting performs well empirically and is
recommended to remove bias induced by overfitting. Moreover,
cross-fitting allows to exploit full efficiency: Every fold is used once
for training the ML methods and once for estimation of the score
\citep[C6]{dml2018}. A short example on the efficiency gains
associated with cross-fitting is provided in Figure~\ref{crossfit}.

\subsubsection{Score function}

Users may want to adjust the score function $\psi(W;\theta_0, \eta_0)$, for example, to adjust the DML estimators in terms of a re-weighting, e.g.,~to adjust for missing outcome via inverse probability of censoring weight (IPCW). An alternative to the choices provided in \pkg{DoubleML} is to pass a function via the argument \code{score} during initialization of the model object. The following examples are equivalent. In the first example, we use the score option \code{"partialling out"} for the PLR model whereas in the second case, we explicitly provide a function that implements the same score. The arguments used in the function refer to the internal objects that implement the theoretical quantities in Equation~\ref{scorepartiallingout}.

Use score \code{"partialling out"}.
\begin{CodeChunk}
\begin{CodeInput}
R> set.seed(314)
R> doubleml_plr_partout <- DoubleMLPLR$new(doubleml_data, ml_l, ml_m,
+    score = "partialling out")
R> doubleml_plr_partout$fit()
R> doubleml_plr_partout$summary()
\end{CodeInput}
\begin{CodeOutput}
Estimates and significance testing of the effect of target variables
\end{CodeOutput}
\end{CodeChunk}

\pagebreak

\begin{CodeChunk}
\begin{CodeOutput}
  Estimate. Std. Error t value Pr(>|t|)    
d    0.5108     0.0959   5.326    1e-07 ***
---
Signif. codes:  0 '***' 0.001 '**' 0.01 '*' 0.05 '.' 0.1 ' ' 1
\end{CodeOutput}
\end{CodeChunk}
We define the function that implements the same score and specify the
argument \code{score} accordingly. The function must return a named
list with entries \code{psi\_a} and \code{psi\_b} to pass values for
computation of the score.

Required input: 
\begin{itemize}
\tightlist
\item \code{y}: dependent variable,
\item \code{d}: treatment variable,
\item \code{l\_hat}: predicted values from regression of $Y$ on $X$,
\item \code{m\_hat}: predicted values from regression of $D$ on $X$,
\item \code{g\_hat}: predicted values from regression of $Y - D \cdot \theta$ on $X$, can be ignored in this example,
\item \code{smpls}: sample split under consideration, can be ignored in this example.
\end{itemize}
\begin{CodeChunk}
\begin{CodeInput}
R> score_manual <- function(y, d, l_hat, m_hat, g_hat, smpls) {
+    resid_y = y - l_hat
+    resid_d = d - m_hat
+    psi_a = -1 * resid_d * resid_d
+    psi_b = resid_d * resid_y
+    psis = list(psi_a = psi_a, psi_b = psi_b)
+    return(psis)
+ }
R> set.seed(314)
R> doubleml_plr_manual <- DoubleMLPLR$new(doubleml_data, ml_l, ml_m,
+    score = score_manual)
R> doubleml_plr_manual$fit()
R> doubleml_plr_manual$summary()
\end{CodeInput}
\begin{CodeOutput}
Estimates and significance testing of the effect of target variables
  Estimate. Std. Error t value Pr(>|t|)    
d    0.5108     0.0959   5.326    1e-07 ***
---
Signif. codes:  0 '***' 0.001 '**' 0.01 '*' 0.05 '.' 0.1 ' ' 1
\end{CodeOutput}
\end{CodeChunk}

\subsection{A short simulation study} \label{simstudy}

To illustrate the validity of the implemented double machine learning
estimators, we perform a brief simulation study.

\begin{figure}[t!]
  \centering
\includegraphics[width=0.8\textwidth, trim=15 20 0 0, clip]{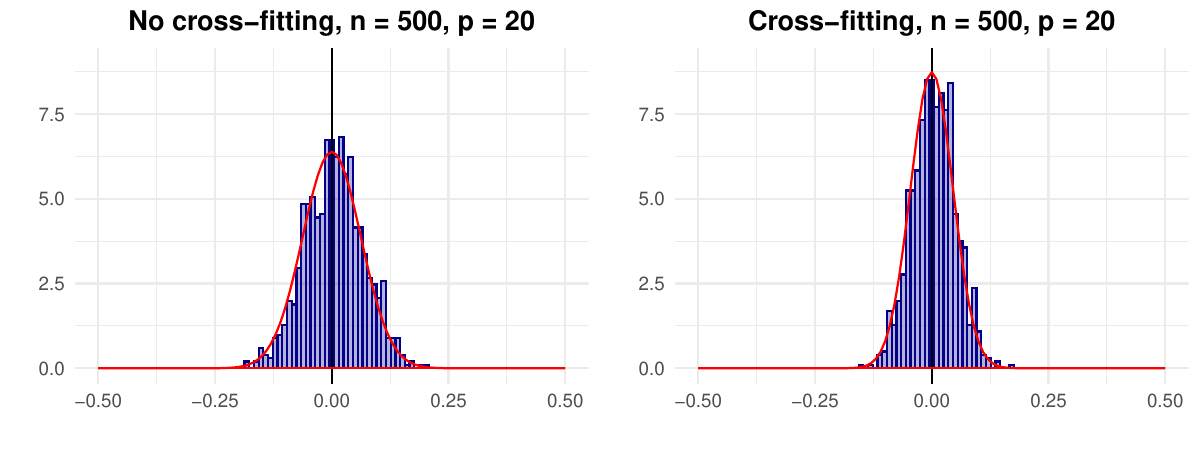}
\caption{%
  Illustration of efficiency gains due to the use of cross-fitting. Left panel:
Histogram of the centered DML estimator without cross-fitting,
$\tilde{\theta}^{\text{nocf}}_0-\theta_0$.  $\hat{\theta}^{\text{nocf}}_0$ is
the double machine learning estimator obtained from a sample split into two
folds. One fold is used for estimation of the nuisance parameters and the second
fold is used for evaluation of the score function and estimation. The empirical
distribution can be well-approximated by a normal distribution as indicated by
the red curve. Right panel: Histogram of the centered DML estimator with
cross-fitting, $\tilde{\theta}_0-\theta_0$. The estimator is obtained from a
split into two folds and application of Algorithm 2 (DML2). In both cases, the
estimators are based on estimation of $g_0$ and $m_0$ with random forests and an
orthogonal score function provided in Equation~\ref{scorepartiallingout}.
Moreover, exactly the same data sets and exactly the same partitions are used
for sample splitting. The empirical distribution of the estimator that is based
on cross-fitting exhibits a more pronounced concentration around zero, which
reflects the smaller standard errors.
}\label{crossfit}
\end{figure}

\subsubsection{The role of cross-fitting}

As mentioned before the use of the
cross-fitting Algorithms 1 (DML1) and 2 (DML2) makes it possible to use
sample splitting and exploit full efficiency at the same time. To
illustrate the superior performance due to cross-fitting, we compare the
double machine learning estimator with and without a cross-fitting
procedure in the simulation setting that was presented in Section
\ref{basicideaplr}. Figure~\ref{crossfit} illustrates that efficiency
gains can be achieved if the role of the random partitions is swapped in
the estimation procedure. Using cross-fitting makes it possible to
obtain smaller standard errors for the DML estimator: The empirical
distribution of the double machine learning estimator that is based on
the cross-fitting Algorithm 2 (DML2) exhibits a more pronounced
concentration around zero.

\subsubsection{Inference on a structural parameter in key causal models}

\begin{figure}[t!]
  \centering
  \includegraphics[width=0.8\textwidth]{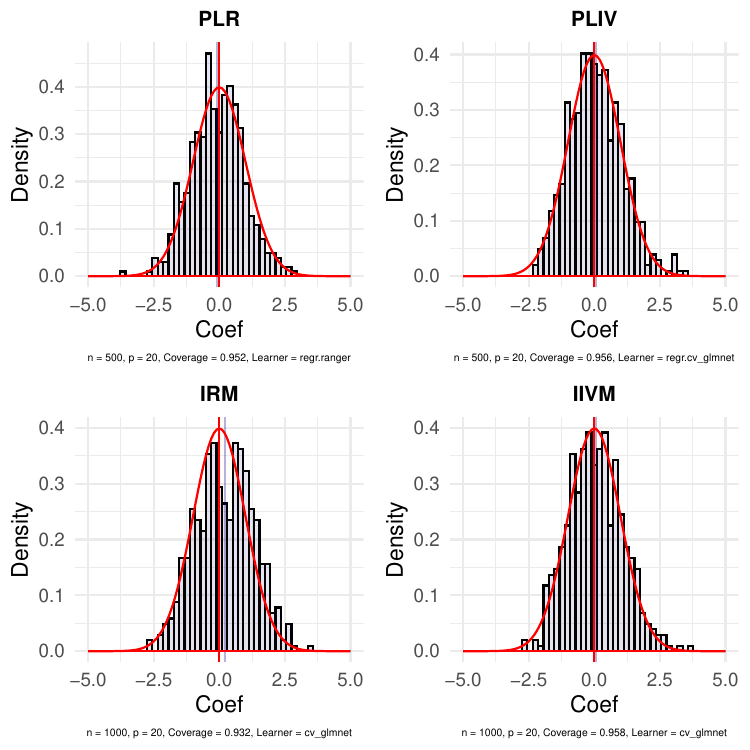}
  \caption{%
    Histogram of double machine learning estimators in key causal models. The
figure shows the histograms of the realizations of the DML estimators in the PLR
(top left), PLIV (top right), IRM (bottom left), and IIVM (bottom right) as
obtained in $R=500$ independent repetitions. Additional information on the data
generating processes and implemented estimators are presented in the main text
and the appendix.
}\label{sim_results}
\end{figure}

We provide simulation results for double machine learning estimators in
the presented key causal models in Figure~\ref{sim_results}. In a
replication of the simulation example in Section~\ref{basicideaplr}, we
show that the confidence intervals for the DML estimator in the
partially linear regression model achieves an empirical coverage (=
\(0.952\)) close to the specified level of \(1-\alpha=0.95\). The
estimator is, again, based on a random forest learner. The corresponding
results are presented in the top-left panel of Figure~\ref{sim_results}.

In a simulated example of a PLIV model, the DML confidence interval that
is based on a lasso learner (\code{regr.cv\_glmnet} of \pkg{mlr3})
achieves a coverage of 95.6\%. The underlying data generating process is
based on a setting considered in \citet{CHSAERpp} with one instrumental
variable. Moreover for simulations of the IRM model, we make use of a
DGP of \citet{belloni2017program}. The DGP for the IIVM is inspired by a
simulation run in \citet{latest}. We present the formal DGPs in the
appendix. To perform estimation of the nuisance functions in the
interactive models, we employ the regression and classification
predictors \code{regr.cv\_glmnet} and \code{classif.cv\_glmnet} as
provided by the \pkg{mlr3} package. In all cases, we employ the
cross-validated \code{lambda.min} choice of the penalty parameter with
five folds, in other words, that \(\lambda\) value that minimizes the
cross-validated mean squared error. Figure~\ref{sim_results} shows that
the empirical distribution of the centered estimators as obtained in
finite sample settings is relatively well-approximated by a normal
distribution. In all models the empirical coverage that is achieved by
the constructed confidence bands is close to the nominal level.

\subsubsection{Simultaneous inference}

To verify the finite-sample performance of the implemented methods for
simultaneous inference, we perform a small simulation study in a
regression setup which is similar as the one used in
\citet{bach2018valid}. We would like to perform valid simultaneous
inference on the coefficients \(\theta\) in the regression model
\begin{align*}
y_i = \beta_0 + d_i^{\top}\theta + \varepsilon_i , \quad \quad i = 1,\ldots, n,
\end{align*}
with \(n=1000\) and \(p_1=42\) regressors. The errors
\(\varepsilon_i\) are normally distributed with
\(\varepsilon_i \sim N(0,\sigma^2)\) and variance \(\sigma^2=3\). The
regressors \(d_i\) are generated by a joint normal distribution
\(d_i \sim N(\mu, \Sigma)\) with \(\mu = \mathbf{0}\) and
\(\Sigma_{j,k} = 0.5^{\lvert j-k\rvert}\). The model is sparse in that only the
first \(s=12\) regressors have a non-zero effect on outcome \(y_i\). The
\(p_1\) coefficients \(\theta_1, \ldots, \theta_{p_1}\) are~generated~as
\begin{align*}
\theta_j= \min\left\{\frac{\theta^{\max}}{j^{a}}, \theta^{\min} \right\},
\end{align*} for \(j=1, \ldots, s\) with \(\theta^{\max}=9\),
\(\theta^{\min}=0.75\), and \(a=0.99\). All other coefficients have
values exactly equal to \(0\). Estimation of the nuisance components has
been performed by using the lasso as provided by
\code{regr.cv\_glmnet} in \pkg{mlr3}.

We report the empirical coverage as achieved by a joint
\((1-\alpha)\)-confidence interval for all \(p_1=42\) coefficients and
the realized family-wise error rate of the implemented $p$~value
adjustments in \(R=500\) repetitions in Table~\ref{siminf_simul}. The
finite sample performance of the Romano-Wolf stepdown procedure that is
based on the multiplier bootstrap as well as the classical Bonferroni
and Bonferroni-Holm correction are evaluated. Table~\ref{siminf_simul}
shows that all methods achieve an empirical FWER close to the specified
level of \(\alpha = 0.1\). In all cases, the double machine learning
estimators reject all 12 false null hypotheses in every repetition.

\begin{table}[t!]
\centering
  \begin{tabular}{@{}lrrrr@{}}
\toprule
 & CI & RW & Bonf. & Holm  \\
\midrule
FWER & 0.08 & 0.11 & 0.08 & 0.10 \\ 
Correct rejections & 12.00 & 12.00 & 12.00 & 12.00 \\
\bottomrule
\end{tabular}
\caption{%
Family-wise error rate (FWER) and average number of correct rejections in a
simulation example. CI: Joint confidence interval, RW: Romano-Wolf stepdown
correction, Bonf.: Bonferroni adjustment, Holm: Bonferroni-Holm correction.
}\label{siminf_simul}
\end{table}

\section{Conclusion} \label{conclusion}

In this paper, we provide an overview on the key ingredients and the
major structure of the double/debiased machine learning framework as
established in \citet{dml2018} together with an overview on a collection
of structural models. Moreover, we introduce the \proglang{R}~package
\pkg{DoubleML} that serves as an implementation of the double machine
learning approach. A brief simulation study provides insights on the
finite sample performance of the double machine learning estimator in
the key causal models.

The structure of \pkg{DoubleML} is intended to be flexible with
regard to the implemented structural models, the resampling scheme, the
machine learning methods and the underlying algorithm, as well as the
Neyman-orthogonal scores considered. By providing the \proglang{R}~package
\pkg{DoubleML} together with its \proglang{Python} twin \citep{DoubleMLpython},
we hope to make double machine learning more accessible to users in
practice. Finally, we would like to encourage users to add new
structural models, scores and functionalities to the package.

\section*{Acknowledgments}

This work was funded by the Deutsche Forschungsgemeinschaft (DFG, German
Research Foundation) -- Project Number 431701914.

\bibliography{v108i03}

\newpage

\begin{appendix}

\section{Computation and infrastructure} \label{computationinfra}

The code in the paper has been executed with \pkg{DoubleML}, version
0.5.3.

The simulation study has been run on a x86\_64, darwin17.0 with macos
Big Sur \ldots{} 10.16 system using \proglang{R} version 4.2.3 (2023-03-15). The following packages have been used for estimation:
  \pkg{DoubleML}, version 0.5.3,
  \pkg{data.table}, version 1.14.6,
  \pkg{mlr3}, version 0.14.1,
  \pkg{mlr3tuning}, version 0.17.2,
  \pkg{mlr3learners}, version 0.5.5,
  \pkg{glmnet}, version 4.1-6,
  \pkg{ranger}, version 0.14.1,
  \pkg{paradox}, version 0.11.0,
  \pkg{foreach}, version 1.5.2.

\section{Suppressed code output}  \label{code}

\subsection[Pennsylvania Reemployment Data, Section 7]{Pennsylvania Reemployment Data, Section~\ref{illustration}}

Load data as \code{data.table}.
\begin{CodeChunk}
\small
\begin{CodeInput}
R> library("DoubleML")
R> dt_bonus <- fetch_bonus(return_type = "data.table")
R> dt_bonus
\end{CodeInput}
\begin{CodeOutput}
      inuidur1 female black othrace dep1 dep2 q2 q3 q4 q5 q6 agelt35 agegt54
   1: 2.890372      0     0       0    0    1  0  0  0  1  0       0       0
   2: 0.000000      0     0       0    0    0  0  0  0  1  0       0       0
   3: 3.295837      0     0       0    0    0  0  0  1  0  0       0       0
   4: 2.197225      0     0       0    0    0  0  1  0  0  0       1       0
   5: 3.295837      0     0       0    1    0  0  0  0  1  0       0       1
  ---                                                                       
5095: 2.302585      0     0       0    0    0  0  1  0  0  0       1       0
5096: 1.386294      0     0       0    0    1  1  0  0  0  0       0       0
5097: 2.197225      0     0       0    0    1  1  0  0  0  0       1       0
5098: 1.386294      0     0       0    0    0  0  0  0  1  0       0       1
5099: 3.295837      0     0       0    0    0  0  0  1  0  0       0       1
      durable lusd husd tg
   1:       0    0    1  0
   2:       0    1    0  0
   3:       0    1    0  0
   4:       0    0    0  1
   5:       1    1    0  0
  ---                     
5095:       0    0    0  1
5096:       0    0    0  1
5097:       0    1    0  0
5098:       0    0    0  1
5099:       1    1    0  0
\end{CodeOutput}
\begin{CodeInput}
R> obj_dml_data_bonus <- DoubleMLData$new(dt_bonus,
+    y_col = "inuidur1", d_cols = "tg",
+    x_cols = c("female", "black", "othrace", "dep1", "dep2", "q2", "q3",
+    "q4", "q5", "q6", "agelt35", "agegt54", "durable", "lusd", "husd"))
\end{CodeInput}
\end{CodeChunk}
Print data backend: Lists main attributes and methods of a `\code{DoubleMLData}'~object.
\begin{CodeChunk}
\small
\begin{CodeInput}
R> obj_dml_data_bonus
\end{CodeInput}
\end{CodeChunk}
Print data set.
\begin{CodeChunk}
\small
\begin{CodeInput}
R> obj_dml_data_bonus$data
\end{CodeInput}
\begin{CodeOutput}
      inuidur1 female black othrace dep1 dep2 q2 q3 q4 q5 q6 agelt35 agegt54
   1: 2.890372      0     0       0    0    1  0  0  0  1  0       0       0
   2: 0.000000      0     0       0    0    0  0  0  0  1  0       0       0
   3: 3.295837      0     0       0    0    0  0  0  1  0  0       0       0
   4: 2.197225      0     0       0    0    0  0  1  0  0  0       1       0
   5: 3.295837      0     0       0    1    0  0  0  0  1  0       0       1
  ---                                                                       
5095: 2.302585      0     0       0    0    0  0  1  0  0  0       1       0
5096: 1.386294      0     0       0    0    1  1  0  0  0  0       0       0
5097: 2.197225      0     0       0    0    1  1  0  0  0  0       1       0
5098: 1.386294      0     0       0    0    0  0  0  0  1  0       0       1
5099: 3.295837      0     0       0    0    0  0  0  1  0  0       0       1
      durable lusd husd tg
   1:       0    0    1  0
   2:       0    1    0  0
   3:       0    1    0  0
   4:       0    0    0  1
   5:       1    1    0  0
  ---                     
5095:       0    0    0  1
5096:       0    0    0  1
5097:       0    1    0  0
5098:       0    0    0  1
5099:       1    1    0  0
\end{CodeOutput}
\begin{CodeInput}
R> learner_classif_m <- lrn("classif.ranger", num.trees = 500,
+    min.node.size = 2, max.depth = 5)
R> doubleml_irm_bonus <- DoubleMLIRM$new(obj_dml_data_bonus, 
+    ml_g = learner_g, ml_m = learner_classif_m, score = "ATE",
+    dml_procedure = "dml1", n_folds = 5, n_rep = 1)
R> doubleml_irm_bonus
\end{CodeInput}
\begin{CodeOutput}
================= DoubleMLIRM Object ==================


------------------ Data summary      ------------------
Outcome variable: inuidur1
Treatment variable(s): tg
Covariates: female, black, othrace, dep1, dep2, q2, q3,
  q4, q5, q6, agelt35, agegt54, durable, lusd, husd
Instrument(s): 
No. Observations: 5099

------------------ Score & algorithm ------------------
Score function: ATE
DML algorithm: dml1

------------------ Machine learner   ------------------
ml_g: regr.ranger
ml_m: classif.ranger

------------------ Resampling        ------------------
No. folds: 5
No. repeated sample splits: 1
Apply cross-fitting: TRUE

------------------ Fit summary       ------------------
\end{CodeOutput}
\end{CodeChunk}

\subsection[Data backend with multiple treatment variables, Section 7.5]{Data backend with multiple treatment variables, Section~\ref{siminf_example}}

\begin{CodeChunk}
\small
\begin{CodeInput}
R> doubleml_data <- double_ml_data_from_data_frame(df, y_col = "y",
+    d_cols = c("X1", "X2", "X3", "X4", "X5", "X6", "X7", "X8", "X9", "X10"))
\end{CodeInput}
\begin{CodeOutput}
Set treatment variable d to X1.
\end{CodeOutput}
\begin{CodeInput}
R> doubleml_data
\end{CodeInput}
\begin{CodeOutput}
================= DoubleMLData Object ==================


------------------ Data summary      ------------------
Outcome variable: y
Treatment variable(s): X1, X2, X3, X4, X5, X6, X7, X8, X9, X10
Covariates: X11, X12, X13, X14, X15, X16, X17, X18, X19, X20, X21, X22, X23,
  X24, X25, X26, X27, X28, X29, X30, X31, X32, X33, X34, X35, X36, X37, X38,
  X39, X40, X41, X42, X43, X44, X45, X46, X47, X48, X49, X50, X51, X52, X53,
  X54, X55, X56, X57, X58, X59, X60, X61, X62, X63, X64, X65, X66, X67, X68,
  X69, X70, X71, X72, X73, X74, X75, X76, X77, X78, X79, X80, X81, X82, X83,
  X84, X85, X86, X87, X88, X89, X90, X91, X92, X93, X94, X95, X96, X97, X98,
  X99, X100
Instrument(s): 
No. Observations: 500
\end{CodeOutput}
\end{CodeChunk}

\subsection[List of externally provided parameters, Section 7.6]{List of externally provided parameters, Section~\ref{learners}}

\begin{CodeChunk}
\small
\begin{CodeInput}
R> str(doubleml_plr$params)
\end{CodeInput}
\begin{CodeOutput}
List of 2
 $ ml_l:List of 10
  ..$ X1 :List of 1
  .. ..$ lambda: num 0.09
  ..$ X2 :List of 1
  .. ..$ lambda: num 0.085
  ..$ X3 : NULL
  ..$ X4 : NULL
  ..$ X5 : NULL
  ..$ X6 : NULL
  ..$ X7 : NULL
  ..$ X8 : NULL
  ..$ X9 : NULL
  ..$ X10: NULL
 $ ml_m:List of 10
  ..$ X1 :List of 1
  .. ..$ lambda: num 0.1
  ..$ X2 :List of 1
  .. ..$ lambda: num 0.095
  ..$ X3 : NULL
  ..$ X4 : NULL
  ..$ X5 : NULL
  ..$ X6 : NULL
  ..$ X7 : NULL
  ..$ X8 : NULL
  ..$ X9 : NULL
  ..$ X10: NULL
\end{CodeOutput}
\end{CodeChunk}

\subsection[List of internally tuned parameters, Section 7.6]{List of internally tuned parameters, Section~\ref{learners}}

\begin{CodeChunk}
\small
\begin{CodeInput}
R> doubleml_plr$tuning_res$X1
\end{CodeInput}
\begin{CodeOutput}
$ml_l
$ml_l[[1]]
$ml_l[[1]]$tuning_result
$ml_l[[1]]$tuning_result[[1]]
$ml_l[[1]]$tuning_result[[1]]$tuning_result
   lambda learner_param_vals  x_domain regr.mse
1:    0.1          <list[2]> <list[1]> 10.53451

$ml_l[[1]]$tuning_result[[1]]$tuning_archive
    lambda regr.mse warnings errors runtime_learners
 1:  0.100 10.53451        0      0             0.10
 2:  0.095 10.60720        0      0             0.05
 3:  0.085 10.76577        0      0             0.05
 4:  0.055 11.32053        0      0             0.05
 5:  0.060 11.21736        0      0             0.05
 6:  0.050 11.42918        0      0             0.09
 7:  0.075 10.93077        0      0             0.05
 8:  0.065 11.11709        0      0             0.05
 9:  0.080 10.84518        0      0             0.03
10:  0.070 11.02168        0      0             0.09
11:  0.090 10.68576        0      0             0.07
                                   uhash  x_domain           timestamp batch_nr
 1: 2407e852-06a7-4756-ace6-42524bc37e34 <list[1]> 2023-01-31 14:49:37        1
 2: 3a35f2c6-b78c-4416-89c9-e69158d2716b <list[1]> 2023-01-31 14:49:37        2
 3: c78c69f3-3a70-4493-afec-121320689918 <list[1]> 2023-01-31 14:49:37        3
 4: 3ffd8bcd-fd2a-46d0-b4b2-c9d0445fba2a <list[1]> 2023-01-31 14:49:37        4
 5: ba275b12-edc5-4c79-8630-4c9c5285095a <list[1]> 2023-01-31 14:49:38        5
 6: ff65786b-19fa-4393-9a9a-4627f07d2f9f <list[1]> 2023-01-31 14:49:38        6
 7: 728bcdef-cfad-4f65-875c-0428f1bc4339 <list[1]> 2023-01-31 14:49:38        7
 8: 96dcc4eb-652a-4825-9481-2fa11b7b274a <list[1]> 2023-01-31 14:49:38        8
 9: e08f6536-0b71-4a82-919f-de35e1872c3f <list[1]> 2023-01-31 14:49:38        9
10: bb30cbc8-8324-441b-bd61-e75d9e22892c <list[1]> 2023-01-31 14:49:39       10
11: 6bac97d1-86cf-43ea-988a-4e7916a13460 <list[1]> 2023-01-31 14:49:39       11

$ml_l[[1]]$tuning_result[[1]]$params
NULL


$ml_l[[1]]$params
$ml_l[[1]]$params[[1]]
$ml_l[[1]]$params[[1]]$family
[1] "gaussian"

$ml_l[[1]]$params[[1]]$lambda
[1] 0.1


$ml_l$params
$ml_l$params[[1]]
$ml_l$params[[1]]$family
[1] "gaussian"

$ml_l$params[[1]]$lambda
[1] 0.1


$ml_m
$ml_m[[1]]
$ml_m[[1]]$tuning_result
$ml_m[[1]]$tuning_result[[1]]
$ml_m[[1]]$tuning_result[[1]]$tuning_result
   lambda learner_param_vals  x_domain  regr.mse
1:    0.1          <list[2]> <list[1]> 0.9794034

$ml_m[[1]]$tuning_result[[1]]$tuning_archive
    lambda  regr.mse warnings errors runtime_learners
 1:  0.090 0.9798230        0      0             0.04
 2:  0.055 0.9971462        0      0             0.07
 3:  0.075 0.9830963        0      0             0.05
 4:  0.050 1.0045139        0      0             0.06
 5:  0.100 0.9794034        0      0             0.06
 6:  0.060 0.9907519        0      0             0.05
 7:  0.065 0.9869171        0      0             0.06
 8:  0.095 0.9797396        0      0             0.06
 9:  0.085 0.9804282        0      0             0.04
10:  0.070 0.9848766        0      0             0.08
11:  0.080 0.9813190        0      0             0.06
                                   uhash  x_domain           timestamp batch_nr
 1: 06cd05b2-2daa-4600-a982-3d35037604a1 <list[1]> 2023-01-31 14:49:39        1
 2: 4aa95e02-b7e3-49a0-b0ae-a5eea57cb686 <list[1]> 2023-01-31 14:49:39        2
 3: 7a97b5c0-44cb-458a-9694-a4205cb8652e <list[1]> 2023-01-31 14:49:40        3
 4: 190a5ba4-00ca-412b-a021-bfe92ba46375 <list[1]> 2023-01-31 14:49:40        4
 5: e83d84d3-91a4-4969-8310-c6e289d3182c <list[1]> 2023-01-31 14:49:40        5
 6: ab176601-1969-4424-8a14-df3c6cf74d84 <list[1]> 2023-01-31 14:49:40        6
 7: 0c3d1269-412b-4734-8fe2-1b9bb7319479 <list[1]> 2023-01-31 14:49:40        7
 8: a808a20c-b862-4665-a83e-85f7d6489536 <list[1]> 2023-01-31 14:49:41        8
 9: f7cdb8f7-eb89-48fd-b295-5dae06f55cf7 <list[1]> 2023-01-31 14:49:41        9
10: 0a503598-db1a-48bd-8d75-6b3620a4a0e2 <list[1]> 2023-01-31 14:49:41       10
11: 47c429d6-7bee-4a77-875b-8d4cdf23f7ee <list[1]> 2023-01-31 14:49:41       11

$ml_m[[1]]$tuning_result[[1]]$params
NULL


$ml_m[[1]]$params
$ml_m[[1]]$params[[1]]
$ml_m[[1]]$params[[1]]$family
[1] "gaussian"

$ml_m[[1]]$params[[1]]$lambda
[1] 0.1


$ml_m$params
$ml_m$params[[1]]
$ml_m$params[[1]]$family
[1] "gaussian"

$ml_m$params[[1]]$lambda
[1] 0.1
\end{CodeOutput}
\end{CodeChunk}
The tuned parameters:
\begin{CodeChunk}
\small
\begin{CodeInput}
R> str(doubleml_plr$params)
\end{CodeInput}
\begin{CodeOutput}
List of 2
 $ ml_l:List of 10
  ..$ X1 :List of 2
  .. ..$ family: chr "gaussian"
  .. ..$ lambda: num 0.1
  ..$ X2 :List of 2
  .. ..$ family: chr "gaussian"
  .. ..$ lambda: num 0.1
  ..$ X3 :List of 2
  .. ..$ family: chr "gaussian"
  .. ..$ lambda: num 0.1
  ..$ X4 :List of 2
  .. ..$ family: chr "gaussian"
  .. ..$ lambda: num 0.09
  ..$ X5 :List of 2
  .. ..$ family: chr "gaussian"
  .. ..$ lambda: num 0.07
  ..$ X6 :List of 2
  .. ..$ family: chr "gaussian"
  .. ..$ lambda: num 0.085
  ..$ X7 :List of 2
  .. ..$ family: chr "gaussian"
  .. ..$ lambda: num 0.085
  ..$ X8 :List of 2
  .. ..$ family: chr "gaussian"
  .. ..$ lambda: num 0.08
  ..$ X9 :List of 2
  .. ..$ family: chr "gaussian"
  .. ..$ lambda: num 0.09
  ..$ X10:List of 2
  .. ..$ family: chr "gaussian"
  .. ..$ lambda: num 0.075
 $ ml_m:List of 10
  ..$ X1 :List of 2
  .. ..$ family: chr "gaussian"
  .. ..$ lambda: num 0.1
  ..$ X2 :List of 2
  .. ..$ family: chr "gaussian"
  .. ..$ lambda: num 0.095
  ..$ X3 :List of 2
  .. ..$ family: chr "gaussian"
  .. ..$ lambda: num 0.095
  ..$ X4 :List of 2
  .. ..$ family: chr "gaussian"
  .. ..$ lambda: num 0.095
  ..$ X5 :List of 2
  .. ..$ family: chr "gaussian"
  .. ..$ lambda: num 0.1
  ..$ X6 :List of 2
  .. ..$ family: chr "gaussian"
  .. ..$ lambda: num 0.1
  ..$ X7 :List of 2
  .. ..$ family: chr "gaussian"
  .. ..$ lambda: num 0.1
  ..$ X8 :List of 2
  .. ..$ family: chr "gaussian"
  .. ..$ lambda: num 0.1
  ..$ X9 :List of 2
  .. ..$ family: chr "gaussian"
  .. ..$ lambda: num 0.1
  ..$ X10:List of 2
  .. ..$ family: chr "gaussian"
  .. ..$ lambda: num 0.1
\end{CodeOutput}
\end{CodeChunk}

\section{Data generating processes, simulation study}

\subsection{Data generating process for PLIV simulation}

The DGP is based on \citet{CHSAERpp} and defined as
\begin{align*}
  \begin{aligned}
    z_i &= \Pi x_i + \zeta_i,\\
    d_i &= x_i^{\top} \gamma + z_i^{\top} \delta + u_i,\\
    y_i &= \alpha d_i + x_i^{\top} \beta + \varepsilon_i,
  \end{aligned}
\end{align*}
with
\begin{align*}
\left(\begin{matrix} \varepsilon_i \\ u_i \\ \zeta_i \\ x_i \end{matrix} \right) \sim \mathcal{N}\left(0, \left(\begin{matrix} 1 & 0.6 & 0 & 0 \\ 0.6 & 1 & 0 & 0 \\ 0 & 0 & 0.25 I_{p_n^z} & 0 \\ 0 & 0 & 0 & \Sigma \end{matrix} \right) \right)
\end{align*}
where \(\Sigma\) is a \(p^x_n \times p^x_n\) matrix with
entries \(\Sigma_{kj} = 0.5^{\lvert k-j\rvert}\) and \(I_{p_n^z}\) is an identity
matrix with dimension \(p_n^z \times p_n^z\). \(\beta=\gamma\) is a
\(p^x_n\)-vector with entries \(\beta=\frac{1}{j^2}\) and
\(\Pi = (I_{p_n^z}, 0_{p_n^z \times (p_n^x - p_n^z)})\). In the
simulation example, we have one instrument, i.e.,~\(p^z_n=1\) and
\(p^x_n=20\) regressors \(x_i\). In the simulation study, data sets with
\(n = 500\) observations are generated in \(R = 500\) independent
repetitions.

\subsection{Data generating process for IRM simulation}

The DGP is based on a simulation study in \citet{belloni2017program} and
defined as
\begin{align*}
\begin{aligned}
d_i &= 1\left\{ \frac{\exp(c_d x_i^{\top} \beta)}{1+\exp(c_d x_i^{\top} \beta)} > v_i \right\}, & &v_i \sim \mathcal{U}(0,1),\\
y_i &= \theta d_i + c_y x_i^{\top} \beta d_i + \zeta_i, & &\zeta_i \sim \mathcal{N}(0,1),
\end{aligned}
\end{align*}
with covariates \(x_i \sim \mathcal{N}(0, \Sigma)\) where
\(\Sigma\) is a matrix with entries \(\Sigma_{kj} = 0.5^{\lvert k-j\rvert}\).
\(\beta\) is a \(p_x\)-dimensional vector with entries
\(\beta_j= \frac{1}{j^2}\) and the constants \(c_y\) and \(c_d\) are
determined as \begin{align*}
c_y = \sqrt{\frac{R_y^2}{(1-R_y^2) \beta^{\top} \Sigma \beta}}, \qquad c_d = \sqrt{\frac{(\pi^2 /3) R_d^2}{(1-R_d^2) \beta^{\top} \Sigma \beta}}.
\end{align*} We set the values of \(R_y^2=0.5\) and \(R_d^2=0.5\) and
consider a setting with \(n=1000\) and \(p=20\). Data generation and
estimation have been performed in \(R = 500\) independent replications.

\subsection{Data generating process for IIVM simulation}

The DGP is defined as
\begin{align*}
  \begin{aligned}
    d_i &= 1\left\lbrace \alpha_x Z + v_i > 0 \right\rbrace,\\
    y_i &= \theta d_i + x_i^{\top} \beta + u_i,
\end{aligned}
\end{align*}
with \(Z\sim \text{Bernoulli}(0.5)\) and
\begin{align*}
\left(\begin{matrix} u_i \\ v_i \end{matrix} \right) \sim \mathcal{N}\left(0, \left(\begin{matrix} 1 & 0.3 \\ 0.3 & 1 \end{matrix} \right) \right).
\end{align*}
The covariates are drawn from a multivariate normal
distribution with \(x_i\sim \mathcal{N}(0, \Sigma)\) with entries of the
matrix \(\Sigma\) being \(\Sigma_{kj} = 0.5^{\lvert j-k\rvert}\) and \(\beta\)
being a \(p_x\)-dimensional vector with \(\beta_j=\frac{1}{\beta^2}\).
The data generating process is inspired by a process used in a
simulation in \citet{latest}. In the simulation study, data sets with
\(n = 1000\) observations and \(p_x = 20\) confounding variables \(x_i\)
have been generated in \(R = 500\) independent repetitions.

\end{appendix}

\end{document}